%% file: kdd2024.tex
\documentclass[sigconf]{acmart}
\makeatletter
\def\@ACM@checkaffil{
    \if@ACM@instpresent\else
    \ClassWarningNoLine{\@classname}{No institution present for an affiliation}%
    \fi
    \if@ACM@citypresent\else
    \ClassWarningNoLine{\@classname}{No city present for an affiliation}%
    \fi
    \if@ACM@countrypresent\else
        \ClassWarningNoLine{\@classname}{No country present for an affiliation}%
    \fi
}
\makeatother

\usepackage{amsmath}
\usepackage{bbm}
\usepackage{xspace}

\usepackage{amssymb}
\usepackage{subfigure}
\usepackage{graphicx}
\usepackage[inline]{enumitem}
\usepackage{multicol}
\usepackage{multirow}
\usepackage{pgf-pie}
\usepackage{algpseudocode}
\usepackage{algorithm}
\usepackage{pifont}
\usepackage{xcolor}
\usepackage{tcolorbox}
\tcbuselibrary{skins, breakable}
\newcommand{\cmark}{\textcolor{green}{\ding{51}}}%
\newcommand{\xmark}{\textcolor{red}{\ding{55}}}%

\input{macro}

\AtBeginDocument{%
  \providecommand\BibTeX{{%
    \normalfont B\kern-0.5em{\scshape i\kern-0.25em b}\kern-0.8em\TeX}}}

\setcopyright{acmlicensed}
\copyrightyear{2018}
\acmYear{2018}
\acmDOI{XXXXXXX.XXXXXXX}
\acmConference[Conference acronym 'XX]{Make sure to enter the correct
  conference title from your rights confirmation emai}{June 03--05,
  2018}{Woodstock, NY}
\acmISBN{978-1-4503-XXXX-X/18/06}

\begin{document}


\title{Understanding the Weakness of Large Language Model Agents within a Complex Android Environment}

\author{Mingzhe Xing}
\authornote{This work is done during the internship at Microsoft Research.}
\affiliation{%
  \institution{Peking University}
  }

\author{Rongkai Zhang}
\affiliation{%
  \institution{Peking University}
  }

\author{Hui Xue}
\affiliation{%
  \institution{Microsoft Research}
  }

\author{Qi Chen}
\affiliation{%
  \institution{Microsoft Research}
  }

\author{Fan Yang}
\affiliation{%
  \institution{Microsoft Research}
  }

\author{Zhen Xiao}
\authornote{Corresponding author.}
\affiliation{%
  \institution{Peking University}
  }


\begin{abstract}
Large language models (LLMs) have empowered intelligent agents to execute intricate tasks within \emph{domain-specific software} such as browsers and games. However, when applied to \emph{general-purpose software systems} like operating systems, LLM agents face three primary challenges. Firstly, the \emph{action space is vast and dynamic}, posing difficulties for LLM agents to maintain an up-to-date understanding and deliver accurate responses. Secondly, real-world tasks often require \emph{inter-application cooperation}, demanding farsighted planning from LLM agents. Thirdly, agents need to identify optimal solutions \emph{aligning with user constraints}, such as security concerns and preferences.
These challenges motivate {\ours}, an environment and benchmark designed to evaluate LLM agents on a modern operating system. To address high-cost of manpower, we design a scalable and semi-automated method to construct the benchmark.
In the task evaluation, \ours incorporates accurate and adaptive metrics to address the issue of non-unique solutions. Our findings reveal that even state-of-the-art LLM agents struggle in cross-APP scenarios and adhering to specific constraints. Additionally, we identify a lack of four key capabilities, \ie understanding, reasoning, exploration, and reflection, as primary reasons for the failure of LLM agents. Furthermore, we provide empirical analysis on the failure of reflection, and improve the success rate by 27\% with our proposed exploration strategy. This work is the first to present valuable insights in understanding fine-grained weakness of LLM agents, and offers a path forward for future research in this area.
Environment, benchmark, and evaluation code for \ours are released at \url{https://github.com/AndroidArenaAgent/AndroidArena}.
\end{abstract}

\maketitle

\input{sections/intro}
\input{sections/relatedwork}
\input{sections/method}

\input{sections/experiment}
\input{sections/conclusion}

\bibliographystyle{ACM-Reference-Format}
\bibliography{ref.bib}

\appendix
\clearpage
\input{sections/appendix}
\end{document}

%% file: macro.tex
\newcommand{\eg}{\textit{e.g.,}\xspace}
\newcommand{\ie}{\textit{i.e.,}\xspace}
\newcommand{\ours}{\texttt{AndroidArena}\xspace}

%% file: sections/intro.tex
\section{Introduction}

Large language models~(LLMs) have shown great potentials in understanding hidden intent from human and commonsense reasoning~\cite{wang2023can}. This makes it possible to utilize \emph{LLM as agent}~\cite{xi2023rise, wang2023survey}, an intelligent entity capable of making decisions and executing actions based upon the perceived state of environment.
An example is for LLMs to interact with \emph{domain-specific software}, such as databases~\cite{li2023can}, games~\cite{wang2023voyager} and browsers~\cite{zhou2023webarena}, for task completion.

More recently, new LLM-based agents have emerged to interact with \emph{general-purpose software systems}, such as operating systems along with their installed APPs, to accomplish more complex \emph{open-domain tasks}~\cite{yan2023gpt,yang2023appagent}. 
These tasks range from simple actions like setting reminders to more intricate activities like financial management and staying connected with loved ones. 
Complex scenarios in operating systems typically manifest the following characteristics: 1) a \textbf{vast and ever-changing action space} due to real-time internet data exchange, APP installations, and upgrades; 2) an increasing demand for \textbf{cross-APP collaboration} as user tasks become more interconnected and multifaceted; and 3) heightened consideration for \textbf{personal interests and security concerns}.

These characteristics motivate us to establish a new environment and comprehensive benchmark to study the boundaries of LLM agent's capability within a complex software system. 
In this paper, we introduce \ours, an environment built on the Android operating system, accompanied by an evaluation benchmark containing annotated ground truth action sequences. 
\ours supports real-time internet data exchange and dynamic APP management, and enables seamless operations across various APPs. 
These features facilitate the evaluation of LLM agents in a \emph{vast and dynamic action space} and \emph{cross-APP scenarios}. 
Additionally, we propose a scalable method for semi-automatically constructing an instruction benchmark, ensuring comprehensive coverage of APP functionalities. 
Our open-source benchmark, informed by the aforementioned characteristics, evaluates tasks not only within a single APP but also complex tasks requiring collaboration across multiple APPs. 
It further considers tasks subject to constraints such as \emph{user preferences and security considerations}.

\input{tabs/benchmark_compare}

Evaluating tasks within a complex operating system is non-trivial~\cite{li2024personal}, primarily due to the fact that the feasible action sequence for a task is often non-unique. This presents a significant challenge to precisely evaluating agents in multi-step decision-making scenarios.
To address this issue, we devise \textbf{adaptive metrics to evaluate task completion accurately}.
The evaluation results reveal that all state-of-the-art~(SOTA) LLM agents fall short in cross-APP scenarios, with a success rate of less than 60\%, and struggle to fully adhere to specific constraints. 
Notably, GPT-3.5~\cite{openai2023gpt} achieves a 6x higher success rate than LLaMA2-70B~\cite{touvron2023llama}. 
Through meticulous case analysis to understand the causes of failure, we identify and abstract \textbf{four key planning capabilities of LLM agents}, inspired by reinforcement learning~(RL)~\cite{sutton1999reinforcement}: \textbf{understanding}, \textbf{reasoning}, \textbf{exploration}, and \textbf{reflection}.
We design metrics to \textbf{measure these fine-grained capabilities, showing improvement directions for LLM agents}. 
LLaMA2 exhibits weaknesses across all dimensions, and even advanced models like GPT-4~\cite{openai2023gpt} are no exemptions, exhibiting weak reflection and exploration abilities. 
Empirical analysis predominantly attributes the weakness in reflection to low-quality trajectories and sparsity in environment feedback. 
Moreover, we find that by integrating historical visited information into the prompt and balancing exploration and exploitation by the agent, the success rate of specific APPs can improve by 27\%, and the exploration performance is enhanced.

In summary, we make the following contributions.

\begin{itemize}[left=0pt]
    \item We open-source \ours, a benchmark based on the Android operating system, to evaluate daily tasks requiring cross-APP collaboration, as well as considerations for constraints such as security. Additionally, our scalable and semi-automated approach reduces the cost of benchmark construction.
    \item Our findings indicate that STOA models underperform in daily tasks and are not ready for direct product integration. We propose fine-grained metrics that reveal failure causes and highlight four areas for future research: understanding, reasoning, reflection, and exploration. Initial analysis show the failure reasons of reflection, and 27\% of improvement when enhancing exploration.
\end{itemize}

%% file: tabs/benchmark_compare.tex
\begin{table*}[tbp]
    \centering
    \caption{The comparison between our \ours benchmark and existing benchmarks.}
    \begin{tabular}{cccccc}
    \toprule
    \multirow{2}{*}{\shortstack{Benchmark}} & \multirow{2}{*}{\shortstack{Online \\ Evaluation}} & \multirow{2}{*}{\shortstack{Realistic \\ Environment}} & \multirow{2}{*}{\shortstack{Scalably \\ Generated}} & \multirow{2}{*}{\shortstack{Collaborative Tasks \\between APPs}}  & \multirow{2}{*}{\shortstack{Tasks with \\Constraints}} \\
    & \\
    \midrule
    MineCraft~\cite{wang2023voyager} & \cmark & \xmark & \xmark & \xmark & \xmark \\
    Mind2Web~\cite{deng2023mind2web} & \xmark & \cmark & \xmark & \xmark & \xmark \\
    AITW~\cite{rawles2023aitw} & \xmark & \cmark & \xmark & \xmark & \xmark \\
    AndroidEnv~\cite{ToyamaEtAl2021AndroidEnv} & \cmark & \cmark & \xmark & \xmark& \xmark  \\
    WebArena~\cite{zhou2023webarena} & \cmark & \cmark & \xmark & \xmark & \xmark \\
    \midrule
    \ours & \cmark & \cmark & \cmark & \cmark & \cmark \\
    \bottomrule    
    \end{tabular}
    \label{tab:comparision}
\end{table*}

%% file: sections/relatedwork.tex
\section{Background}

\subsection{Frameworks of LLM Agent}
With the emergence of LLMs, the study of LLM agents has begun to thrive.
Early research work~\cite{huang2022language,ahn2022can,qian2023communicative} prompt LLMs to directly generate actions based on environment observations.
ReAct~\cite{yao2022react} is a pioneer work to integrate reasoning and acting in LLM for general task solving. 
It first generates reasoning traces based on history context, subsequently producing actions to interact with the environment. 
Building upon this task-solving paradigm, subsequent agents have been proposed to enhance capabilities in various dimensions.
Reflexion~\cite{shinn2023reflexion} summarizes textual feedback from the environment and then incorporates it as additional context for the LLM agent in the subsequent episode. 
The self-reflective context acts as a semantic gradient signal, offering the agent a concrete direction to improve upon, and facilitates the learning process from prior mistakes for enhancing task performance.
This paper focuses on evaluating the abilities of LLM agents and understanding their weaknesses. We adopt ReAct as the basic agent strategy and Reflexion as an approach to assess the agent's ability of self-reflection.

\subsection{Existing Operating System Task Benchmark} 
Operating Systems~(OS) serve as crucial environments with which humans interact daily, and numerous benchmarks have emerged to evaluate the performances of agents within OS. 
AITW~\cite{rawles2023aitw} stands out as a static image dataset that offers human demonstrations of device interactions. However, the static nature of AITW prevents agents from obtaining a reproducible environment.
On the other hand, AndroidEnv~\cite{ToyamaEtAl2021AndroidEnv} provides support for dynamic interactions with APP. Despite this, it only supports single APP's interaction in each environment instance, which limits its capability to evaluate complex and realistic tasks.
WebArena~\cite{zhou2023webarena} creates tasks simulating human behavior on web browsers. However, it is also limited to automate tasks on a single website, which poses a constraint on its applicability.
While these works have been a source of inspiration, they also highlight the significant challenges of evaluating tasks on OS. Tasks performed by real-world users are often more complex and demanding, requiring the collaboration of multiple APPs. 
Additionally, agents need to consider various constraints such as security and user preference.
Therefore we propose \ours, a reproducible mobile environment that allows for cross-APP access. Alongside this, we introduce a new dataset that encapsulates the richness, difficulty, and constraints of instructions. The detail comparison with other works is listed in Table \ref{tab:comparision}.

\subsection{Existing Metrics for Multi-step Decision}
To assess the performances of LLM agents, a range of metrics are proposed. 
SmartPlay~\cite{wu2023smartplay} and LASER~\cite{ma2023laser} employ success rate and reward to evaluate task completion. 
However, these metrics cannot reflect the detailed completion within individual tasks. 
TPTU~\cite{ruan2023tptu} and Mind2Web~\cite{deng2023mind2web} incorporate a step-wise action alignment method, offering a more nuanced analysis of task completion.
Nevertheless, when applied to multi-step decision-making scenarios where the feasible action sequence for completing a task is not unique, this kind of step-wise matching method may introduce inaccuracies.
In this paper, we propose an adaptive way to accurately assess the task completion, and a set of fine-grained ability evaluation metrics to understand the weaknesses of agents, providing valuable insights into improvement directions for LLM agents.

%% file: sections/method.tex
\section{AndroidArena Environment}
In this section, we introduce the \ours environment, distinguished by its vast and dynamic action space, along with its capability to facilitate cross-APP and constrained task execution.
We begin by offering a formal definition of the mobile task automation process, followed by an overview of the system implementation.
Subsequently, we explore the intricacies of the action space, highlighting its dynamic and expansive nature.
\begin{figure}[tbp]
\centering
    \subfigure[APP screenshot.]{
    \label{fig:subfig:contact_screenshot}
    \includegraphics[width=0.3\columnwidth]{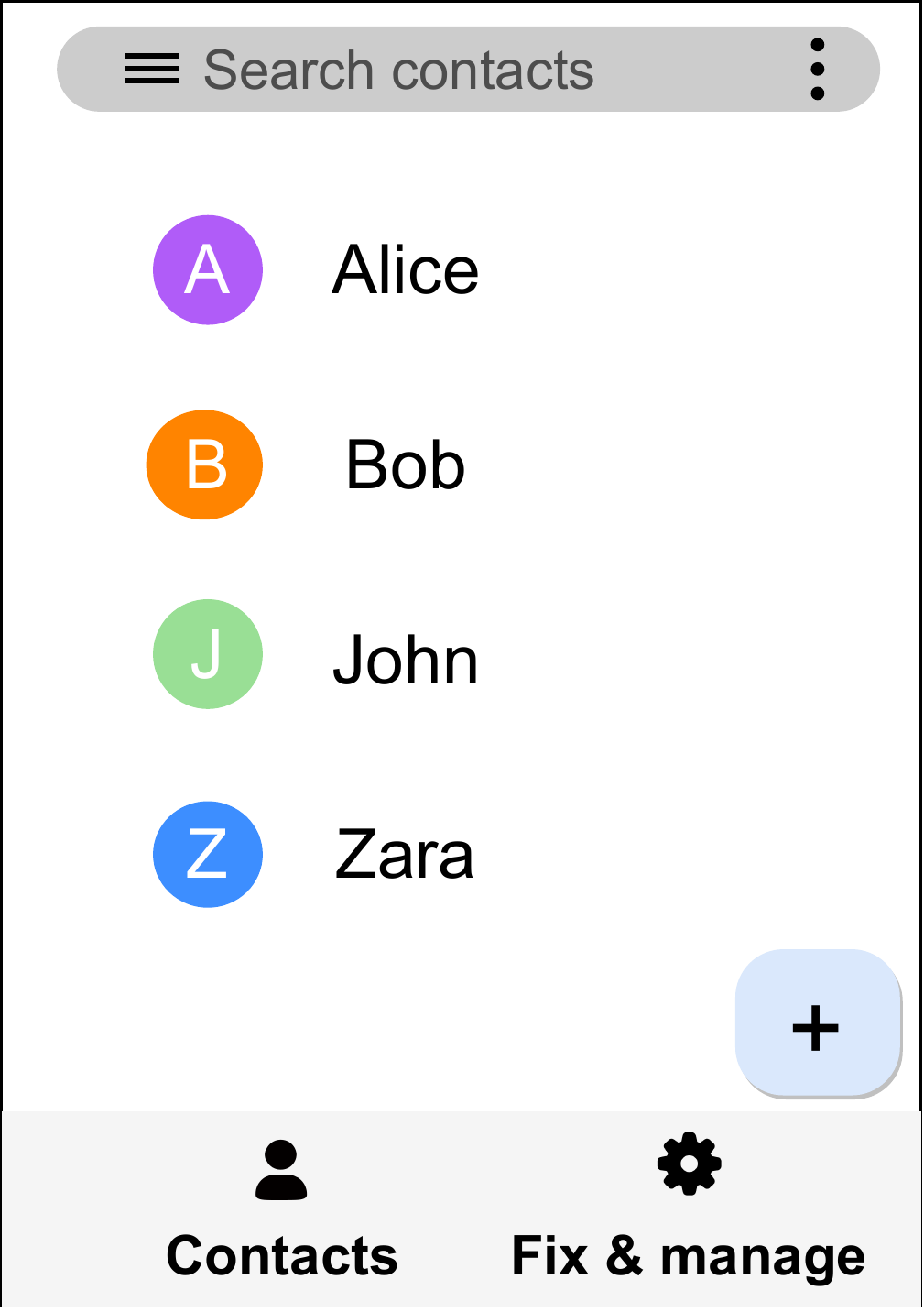}}
    \subfigure[Compressed observation.]{
    \label{fig:subfig:compressed_obs}
    \includegraphics[width=0.6\columnwidth, trim=30 510 140 10, clip]{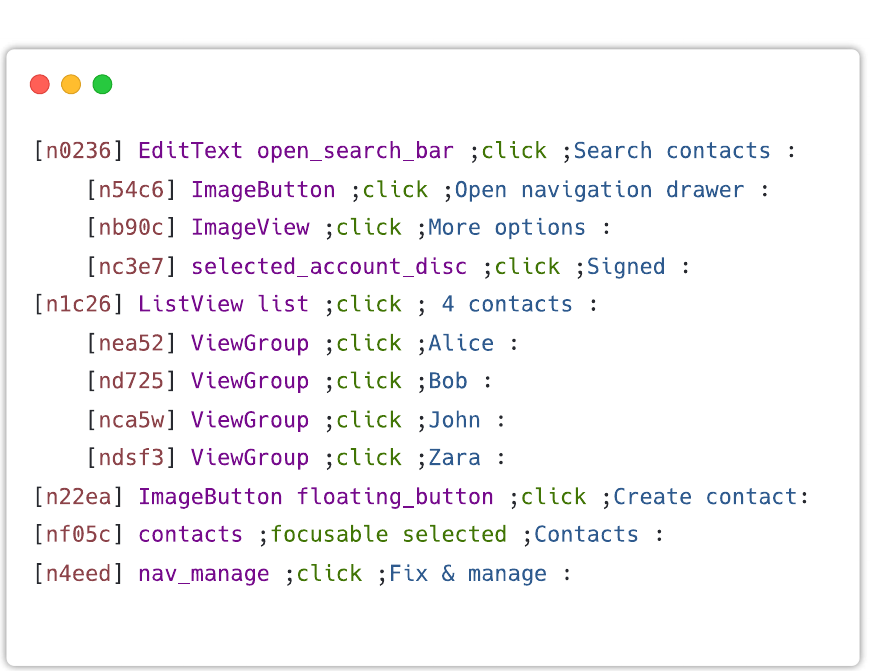}}
\caption{An example of the Contacts APP page and its corresponding compressed observation.}
\label{fig:obs_example}
\end{figure}
\subsection{LLM Agent for Mobile Task Automation}
Given a task presented with a user instruction in natural language, the agent is responsible for making action decisions to complete this instruction on the phone.
This process can be formulated as a Contextual Markov Decision Process (CMDP)~\cite{hallak2015contextual} $\langle \mathcal{C}, \mathcal{S}, \mathcal{A}, \mathcal{T} , r \rangle$.
Context $c\in\mathcal{C}$ is the mobile task explicitly expressed as a textual instruction.
State $s\in\mathcal{S}$ is the current observed phone state, \ie the displayed content on the screen.
Action $a\in\mathcal{A}$ can be performed on the current phone screen, \eg clicks or typing.
Transition function $\mathcal{T}(s^\prime | s, a)$ represents the change in the phone on performing an action.
Reward $r$ is awarded for successful completion of the task.

\textbf{Implementation.}
Our implementation is based on UIAutomator~\cite{gunasekaran2015survey}, a UI testing framework that enables direct operations on UI components. 
With UIAutomator, we offer flexible configurations to render APP page content (\ie the observation space) in two modes: 1) the phone screenshot, a pixel-based representation as perceived by humans, and 2) the textual XML description of the phone screen~(depicted in Fig.~\ref{fig:xml_example} in \S\ref{sec:detailed_compress_method}). 
It is important to note that, given the focus of this work on LLM agents, we exclusively utilize the text modality, while acknowledging that our implementation is capable of supporting multi-modal models.
Each UI component in the screen corresponds to an XML entry, containing its role~(\eg a button), text content, and properties (\eg if clickable) information.
A statistic conducted on eight popular APPs indicates an average token count for the XML exceeding 10,000. 
Consequently, directly feeding the XML into the LLM is impractical due to context length limitations. 
To address this challenge, we propose a two-stage heuristic compression method, involving the removal of decision-irrelevant XML tags and the merging of non-visible or non-functional nodes (the detailed algorithm is provided in \S\ref{sec:detailed_compress_method}) to compress the XML.
As illustrated in Fig.~\ref{fig:subfig:compressed_obs}, the compressed observation maintains the hierarchical structure of the original XML, enabling the LLM agent to comprehend the UI layout via text. Subsequent to compression, each entry is assigned a unique ID (\eg {[nd725]}), facilitating agents in locating the UI element. Our proposed method achieves a compression ratio of 86.6\% across several tested APPs (please see Table~\ref{tab:obs_len} in \S\ref{sec:detailed_compress_method}).
Motivated by previous research~\cite{kwon2023reward} showing superior performance by regarding LLM as reward functions, we employ GPT-4 to quantify the reward $r$, and validate its effectiveness through experiments in \S\ref{sec:macro_eval}.

\subsection{Vast and Dynamic Action Space}

Unlike prior environments~\cite{ToyamaEtAl2021AndroidEnv,zhou2023webarena} focusing on a single APP and only supporting specific actions, our action space is {vast and dynamic}.
It is attributed to the fact that a typical APP may feature hundreds of UI elements available for manipulation, and these UI components exhibit variability owing to real-time internet data exchange.
The vast and dynamic natures are further amplified when considering all the APPs within \ours.
Our designed actions can be categorized into four groups: 1) {APP-level actions} are responsible for installing, launching, and stopping APPs;
2) {Component-level actions} directly operate the UI components such as clicking, typing, and swiping etc;
3) {System-level actions} include turning the screen on and off, adjusting the volume, and taking screenshots etc;
and 4) {Task-level action} is issued when the agent deems the task should finish. 
The complete action space is in \S\ref{sec:Detailed_action_space}.

\section{Scalable Mobile Task Generator}
\begin{figure}[tbp]
    \centering
    \includegraphics[width=0.9\columnwidth]{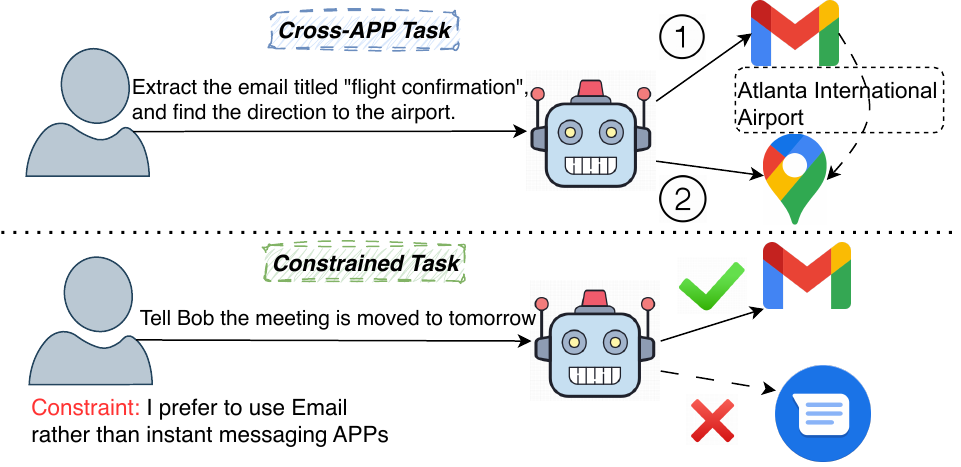}
    \caption{Examples of cross-APP and constrained tasks.}
    \label{fig:task_example}
\end{figure}
\begin{figure*}[tbp]
    \centering
    \includegraphics[width=0.8\textwidth]{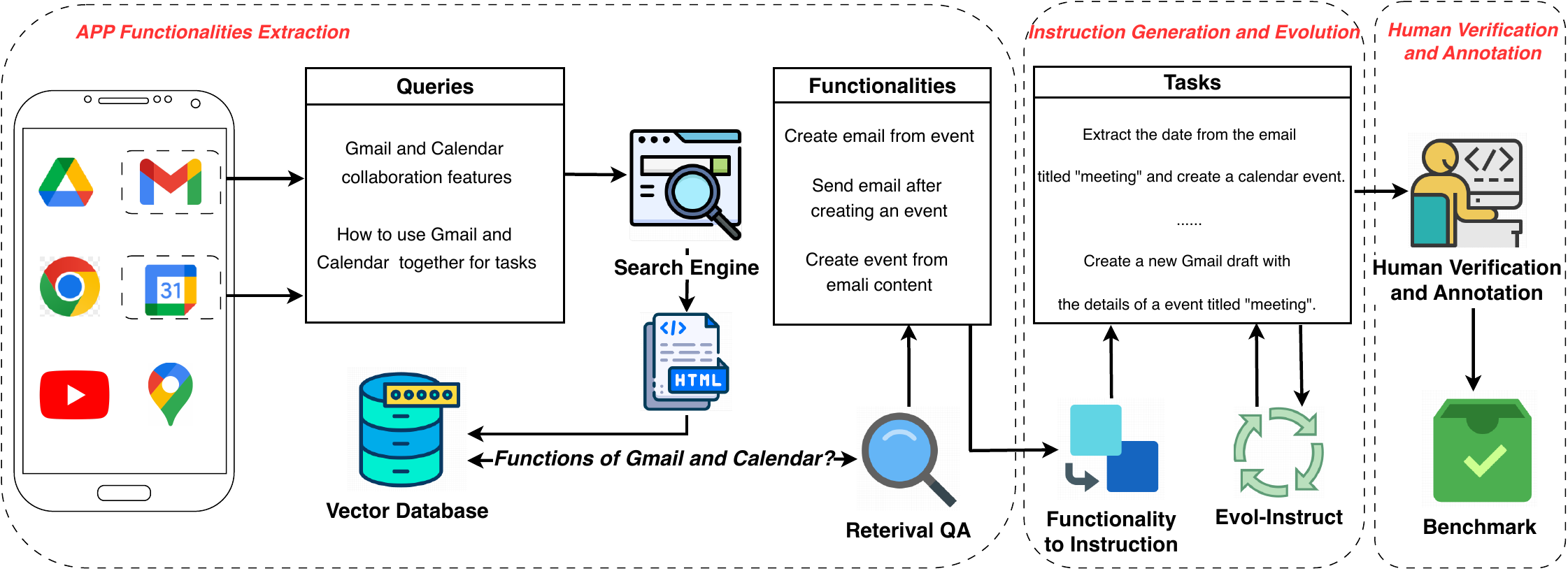}
    \caption{Illustrative example of the MTG workflow for cross-APP~(\ie Gmail and Contacts) tasks construction procedure. The single-APP tasks are generated with the same process but with different query templates and LLM prompts~(please see \S\ref{sec:f2i_prompt}).}
    \label{fig:benchmark_agent}
\end{figure*}

The tasks executed in \ours environment are distinguished from other benchmarks by incorporating {cross-APP collaboration} and {constrained tasks} scenarios, commonly encountered in real-life but ignored in existing benchmarks. 
Even for single-APP tasks, existing benchmarks are either small-scale~\cite{yang2023appagent} or derived from the PixelHelp forum~\cite{rawles2023aitw, yan2023gpt}, a platform dedicated to discussing phone-related issues, thus deviating from routine tasks.
These shortcomings underscore the necessity for a benchmark that exhibits higher \textbf{scalability} and \textbf{aligns closely with human experiences}, while accounting for \textbf{cross-APP} and \textbf{constrained tasks}.

This section outlines our proposed Mobile Task Generator~(MTG in short), a framework for scalable task construction.
MTG not only aligns with typical human interaction patterns, but also encompasses a diverse array of APP functions, enabling to evaluate the agents across a broader spectrum.
The constructed benchmark comprises three task categories: \textit{single-APP tasks}, \textit{cross-APP tasks}, and \textit{constrained tasks}.
The single-APP and cross-APP tasks are crafted to assess the agents' proficiency in solving general tasks, and more complex tasks requiring cooperation between two APPs.
In contrast to the former two categories focusing on task completion, the constrained tasks are designed to evaluate agents' proficiency in comprehending predefined constraints.
In our benchmark, each task consists of a natural language instruction and a sequence of labeled actions for task completion. Constrained tasks additionally include a field of constraints represented in natural language. 
Examples of cross-APP and constrained tasks are shown in Fig.~\ref{fig:task_example}.
The statistical information of our benchmark is presented in Table~\ref{tab:dataset}.
\input{tabs/dataset}

\subsection{Single- and Cross-APP Tasks Construction}
\textbf{APP Functionalities Extraction.}
We incorporate 13 testing APPs from pre-installed Google suite that are designed to work seamlessly with the Android OS and provide essential services.
The complete APP list can be found in \S\ref{sec:benchmark_stat}.
Our objective is to formulate the task instructions that cover rich and diverse functionalities of APPs while aligning with the authentic usage behavior of humans.
To achieve this goal, we propose leveraging insights gleaned from human discussions and shared experiences regarding APPs available on the internet.
Concretely, we first formulate queries centered on the usage of specific APPs and employ search engines to retrieve related webpages.
As depicted in Fig.~\ref{fig:benchmark_agent}, exemplified constructed queries are \textit{``Gmail and Calendar collaboration features''} and \textit{``How to use Gmail and Calendar together for tasks''}. 
We then build a vector database to store these high volume of webpages containing rich functionalities that genuinely engage and concern users.
By retrieving from the database with LLM and a specific prompt~(\S\ref{sec:query_template}), we can extract confined APP functionalities.

\textbf{Instruction Generation and Evolution}
Our next step involves utilizing a LLM with a functionality-to-instruction prompt~(provided in \S\ref{sec:f2i_prompt}) to generate initial task instructions grounded in the identified APP functionalities.
To automatically mass-produce more instructions, we employ the Evol-Instruct~\cite{xu2023wizardlm} strategy to expand the original instructions.
In the application of this strategy, each evolutionary iteration involves using LLM along with two prompts, namely \emph{in-depth evolving} and \emph{in-breadth evolving}.
The in-depth evolving prompt encourages LLM to rewrite instructions by making them more complex and challenging, while in-breadth evolving prompt aims to enhance the feature coverage and overall dataset diversity.
Through the iterative execution of multiple evolutions, we sequentially derive evolution datasets, thereby expanding and refining the pool of task instructions.

\textbf{Human Verification and Annotation}
To construct the benchmark, we engage annotators proficient in operating the testing APPs.
They are first instructed to discern and filter tasks exhibiting repetitiveness, ambiguity, or impossibility to complete.
Subsequently, they document their interactions with the phone.
Given that there might be multiple feasible action sequences for completing a task, they are encouraged to opt for the most concise action plan with the shortest action sequence. 
After completing a task, annotators re-execute the annotated action sequence with a replay script, enabling them easily to verify the accuracy of annotated action sequence. 
Subsequently, the compiled task instructions and action demonstrations are collected into the benchmark dataset.

\vspace{-10pt}
\subsection{Constrained Tasks Construction}
In the context of real-world mobile tasks, often confined by specific user preferences or security considerations, we introduce a constrained task set to assess the agents' capability to comprehend user-defined constraints and make decisions adeptly to avoid violations. 
Specifically, we consider three types of constraints: \emph{APP-level}, \emph{page-level} and \emph{component-level} constraints. 
APP-level constraints involve the preferences of using specific APPs, exemplified by constraints like \textit{``preferring not to use instant message for communication''}. 
Page-level constraints restrict access to a specific page, as seen in scenarios such as \textit{``refraining from entering the label list page in Gmail due to the presence of sensitive information''}. 
Component-level constraints identify specific UI components as sensitive actions, \eg \textit{``do not click the payment button''}.
It is noteworthy that the constrained tasks are meticulously selected from the single-APP task set and manually labeled with natural language constraints along with the corresponding correct action sequences.

\section{Evaluation Metrics}
Designing precise metrics is essential for accurately and comprehensively evaluating agent's performance. 
However, existing metrics employed in multi-step decision-making scenarios~\cite{ruan2023tptu,deng2023mind2web} exhibit \textbf{imprecise} and \textbf{surface-level} evaluation issues, which hinder them to fully understand the performance and weakness of LLM agent.
To address these limitations, we propose a novel set of metrics to evaluate agent performance in a more \textbf{adaptive and precise manner}, and to assess \textbf{fine-grained agent planning abilities}.

\subsection{Adaptive and Precise Task Completion Evaluation}
\label{sec:macro}
To begin with, we introduce the notations of action sequences. 
Given a task, its annotated action sequence can be represented as $\mathbf{a}$ of length $L$, and the actual executed actions is $\mathbf{\hat{a}}$ of length $\hat{L}$.
Exemplary instances of $\mathbf{a}$ and $\mathbf{\hat{a}}$ are illustrated as follows:
\begin{align}
    \mathbf{a}=&\textcolor{red}{AB}CD\textcolor{red}{EF}\textcolor{red}{G} \label{eq:labeled_as}\\
    \mathbf{\hat{a}}=&\textcolor{red}{A}XY\textcolor{red}{B}UVW\textcolor{red}{EF}FF\textcolor{red}{G}Z \label{eq:acutal_as},
\end{align}
where each uppercase character denotes a distinct type of action.
Many existing metrics~\cite{ruan2023tptu, deng2023mind2web} adopt the \emph{step-wise matching} method, which is imprecise in this scenario.
In Eq.~\ref{eq:acutal_as}, the agent identifies the correct action $B$ after two steps of exploration (\ie $X$ and $Y$).
Despite this action sequence not aligning with the ground truth (\ie Eq.~\ref{eq:labeled_as}) in the step-wise manner, it leads to the correct subsequent step and constitutes a valid action sequence for completing the task.
Therefore, previous metrics exhibit inaccuracies in the multi-step decision-making environments where multiple feasible action sequences exist.
In contrast to previous greedy step-wise matching, we propose to align the two sequences in an adaptive way, \ie calculating their \textbf{longest common subsequence}~(LCS) $\mathbf{a}_{lcs}$ (marked in red in Eq.~\ref{eq:labeled_as} and \ref{eq:acutal_as}).
The LCS \textbf{accurately} and \textbf{adaptively} reflects task completion in the multi-step decision-making scenario. 
Based on the accurate LCS, we propose our metrics to evaluate the task completion as follows:
\begin{itemize}[left=0pt]
    \item \textbf{Task Reward~(TR).}
$
    TR=\sum_{i=0}^{L}\gamma^{(L-i)}\mathbbm{1}_i,
$
where $\gamma \in [0, 1]$ is the reward discount factor, $\gamma^{(L-i)}$ assigns higher rewards to the actions that are closer to the final action~(\eg $G$ in Eq.~\ref{eq:labeled_as}), and $\mathbbm{1}_i$ equals 1 when the $i$-th action is in the LCS.
This metric considers both the action matching and the distance towards task success.
\item \textbf{Task Completion Ratio~(TCR).}
$
    TCR=k/L,
$
where $k$ is the index of the last matched action in the LCS. 
This metrics measure the progress of task completion.
\item \textbf{Reversed Redundancy Ratio~(RRR).}
$
    RRR=L/\hat{L}. 
$
It can be used to evaluate the efficiency of the agent completing a task. 
We inverse it for the convenience of comparison, \ie the higher this metric, the greater the efficiency of the agent.
\item \textbf{Success Rate~(SR).}
Unlike the above three metrics relying on ground truth action sequence, the SR is judged by the GPT-4 solely given the trajectory including historical actions and observations.
SR equals $1$ when GPT-4 perceives that the task has been successfully completed, and $0$ when the task is deemed unsuccessful.
This metric is devised for the unsupervised scenario, enhancing the scalability of the evaluation. 
In \S\ref{sec:macro_eval}, we provide statistical evidence to demonstrate the accuracy of SR.
\end{itemize}

\subsection{Understand Root Cause with Fine-grained Abilities Evaluation}
\label{sec:micro_metric}
\input{sections/rl_algo}
In addition to providing adaptive and accurate metrics for evaluating task completion in complex decision-making scenarios, another primary objective of our study is to investigate the underlying root cause contributing to the success or failure of agents planning.
Recognizing that RL serves as a classical and effective approach to address the CMDP problem~\cite{hallak2015contextual}, we abstract the fundamental elements and mechanisms in RL agents, and propose fine-grained capabilities tailored to assess LLM agents.
Here we use the DQN~\cite{mnih2013playing}~(Algorithm~\ref{alg}), one of the most classical RL algorithms, as an example.
We decompose it into four key dimensions, \ie
{\textbf{understanding}}, {\textbf{reasoning}}, {\textbf{exploration}}, and {\textbf{reflection}}.

\textbf{Understanding.} The aspect of {understanding} encompasses the agent's proficiency in comprehending observation and adhering to the action format and space specified in the prompt. 
Unlike RL agents confining their output actions strictly within a predefined action space, the output space of LLM spans the entire vocabulary. 
This imposes a great demand on LLM agents to fully understand and adhere to the specified action format and space. 
Additionally, constrained by phone screen size limit, vital information such as the succinct status description of a checkbox, poses challenges for LLM agents in understanding crucial but brief observed details. 
Consequently, to comprehensively gauge the agent's understanding ability, we formulate three metrics: 
\begin{itemize}[left=0pt]
    \item \textbf{Invalid Format.} The ratio of outputting actions that deviate from the format predefined in prompt.
    \item \textbf{Invalid Action.} The ratio of outputting actions outside the action space specified in prompt.
    \item \textbf{Nuggets Mining.} The ratio of the target element length to the entire observation, assessing the agent's capacity to understand the task context and identify pivotal pieces of information.
    For example, when the agent correctly selects the \textit{Bob} as shown in Fig.~\ref{fig:obs_example}, the Nuggets Mining can be computed as the division of the length of the \textit{[nd725]} entry by the total length of the observation.
\end{itemize}
\input{tabs/macro_evals}
\textbf{Reasoning.} This dimension indicates the agent's capacity to deduce the most suitable action based on the current observation. 
To assess it, two metrics are employed:
\begin{itemize}[left=0pt]
    \item \textbf{Operation Logic.} The inverse number of incorrect actions attempted before successfully finding the correct action.
    Consider Eq.~\ref{eq:labeled_as} and \ref{eq:acutal_as} as an example.
    The agent correctly executes action \textit{B} after two erroneous attempts, \ie $X$ and $Y$.
    Therefore, the Operation Logic for this subsequence is calculated as $1/2$.
    \item \textbf{Awareness of Completion.} The ratio of cases that the agent correctly finds the task completed and issues a finish action.
\end{itemize}
\textbf{Exploration.} LLM agents make decisions from pretraining-derived prior knowledge. 
Due to the static nature of their prior knowledge, certain LLM agents exhibit a proclivity to iteratively execute the same erroneous action~\cite{zhou2023webarena}.
It precludes them from exploring alternative action pathways to ascertain the correct execution path. 
This phenomenon reflects the agent's {exploration} ability, which we quantify by counting the instances of action repetition.
\begin{itemize}[left=0pt]
    \item \textbf{Repeat Actions.}  The ratio of actions resulting in repetitive or cyclical patterns.
\end{itemize}
\textbf{Reflection.} Similar to RL agents, the LLM agents are proven to have the capability to extract insights from previous trials and leverage the insights for subsequent executions~\cite{shinn2023reflexion}.
We utilize the {Reflexion} mechanism to gauge the agent's proficiency in extracting pertinent experiences and applying them judiciously.
\begin{itemize}[left=0pt]
    \item \textbf{Reflexion@K.} 
$
    Reflexion@K=\sum_{i=1}^{K}(SR_i-SR_{i-1}), 
$
where $K$ is the number of Reflexion iterations.
It measures the differences between the original trail and the trail after Reflexion.
\end{itemize}
\textbf{Remark:} The four dimensions are not mutually independent. 
For instance, a prerequisite for reasoning the optimal action is a thorough understanding of the environment and observation.  
Our objective is to assess agent abilities from diverse perspectives rather than segregating them into independent components. 
It is worth noting that our proposed dimensions and metrics can be generalized to other LLM agents, enabling the evaluation of their capabilities in different environments.
The dimension scores are computed as the average of their corresponding metrics~(details provided in \S\ref{sec:micro_compute}).

%% file: tabs/dataset.tex
\begin{table}[tbp]
\centering
\caption{The statistics of our benchmark.}
\label{tab:dataset}
\begin{tabular}{ccc}
\hline
Task Type                       & \#Tasks & Avg. Len. of Action Sequence \\ \hline
single-APP tasks                    & 164                       & 6.13                                     \\
cross-APP tasks                       & 22                        & 11.14                                    \\
constrained tasks & 35                        & 6.03                                     \\ \hline
\end{tabular}
\end{table}

%% file: sections/rl_algo.tex
\begin{algorithm}[tbp]
\begin{algorithmic}
\State Initialize replay memory $\mathcal{D}$ and action-value function $Q$ 
\For{episode $=1,M$} 
\State Initialise state $\phi_1 = \phi(s_1)$
\For {$t=1,T$}
	\State With probability $\epsilon$ select a random action $a_t$     \textcolor{red}{\Comment{\underline{Explore} the environment}}
	\State otherwise select $a_t = \max_{a} Q^*(\phi(s_t), a; \theta)$ \textcolor{red}{\Comment{\underline{Reason} the next action}}
	\State Execute action $a_t$ in emulator and observe reward $r_t$ and state $s_{t+1}$
	\State Preprocess observation $\phi_{t+1} = \phi(s_{t+1})$ \textcolor{red}{\Comment{\underline{Understand} the environment and observation}}
	\State Store transition $\left(\phi_t,a_t,r_t,\phi_{t+1}\right)$ in $\mathcal{D}$
	\State Optimize Q based on a minibatch sampled from $\mathcal{D}$ \textcolor{red}{\Comment{\underline{Reflection} from experience}}
\EndFor
\EndFor
\end{algorithmic}
\caption{Deep Q-learning}
\label{alg}
\end{algorithm}

%% file: tabs/macro_evals.tex
\begin{table*}[tbp]
    \centering
	\begin{minipage}{0.57\textwidth}
		\centering
\caption{Performances evaluated on single-APP and cross-APP tasks. Cross-APP tasks pose a significant challenge for SOTA agents, and highlight a substantial disparity between GPT-4 and other agents.}
\label{tab:macro}
\begin{tabular}{c|cccc|cccc}
\hline
\multicolumn{1}{c|}{} & \multicolumn{4}{c|}{Single-APP Tasks} & \multicolumn{4}{c}{Cross-APP Tasks} \\ \hline
Model                & TR      & TCR     & RRR     & SR     & TR      & TCR     & RRR    & SR     \\ \hline
LLaMA2-13B           & 0.025   & 0.038   & 0.007   & 0.023  & 0.027   & 0.084   & 0.000  & 0.000  \\
LLaMA2-70B           & 0.237   & 0.301   & 0.047   & 0.127  & 0.062   & 0.089   & 0.000  & 0.000  \\
GPT-3.5              & 0.413   & 0.555   & 0.262   & 0.449  & 0.214   & 0.390   & 0.021  & 0.048  \\
GPT-4                & \textbf{0.502}   & \textbf{0.689}   & \textbf{0.755}   & \textbf{0.759}  & \textbf{0.421}   & \textbf{0.746}   & \textbf{0.685}  & \textbf{0.571}  \\ \hline
\end{tabular}
	\end{minipage}
	\hfill
	\begin{minipage}{0.37\textwidth}
		\centering
\caption{The Pearson Correlation Coefficient of SR with information richness~(IR), and with operation complexity~(OC), and with the multiplication of IR and OC.}
\label{tab:app_iroc}
\begin{tabular}{ccc}
\hline
Metrics                     & GPT-3.5 & GPT-4 \\ \hline
IR                          & 0.37                        & 0.62                      \\
OC                          & 0.61                        & 0.28                      \\ 
IR $\times$ OC & 0.68                        & 0.57                      \\ \hline
\end{tabular}
	    \end{minipage}
\end{table*}

%% file: sections/experiment.tex
\section{Experiments and Findings}
In this section, we setup the experiments, and present the experimental results.
We summarize noteworthy findings as follows.
First, existing SOTA agents still exhibit \textbf{substantial room for improvement}~(\S\ref{sec:macro_eval}). 
Second, in contrast to the results observed in prior benchmarks~\cite{liu2023agentbench,crispino2023agent}, \textbf{LLaMA2-70B exhibits inferior planning abilities across various dimensions. GPT-4, while advanced, requires further improvement in the exploration and reflection dimensions.}~(\S\ref{sec:micro_eval}).
\subsection{Evaluation Setting}
We conduct experiments on SOTA open-source and closed-source LLMs. 
The detailed experiment settings are introduced as follows.

\noindent \textbf{Agent Models.} The selected LLMs encompass GPT-\{3.5-turbo, 4\}~\cite{openai2023gpt}, LLaMA2-\{13B-chat, 70B-chat\}~\cite{touvron2023llama}, representing two powerful closed-source and open-source LLM model families, respectively.
Regarding the prompt settings for LLM agents, please refer to \S\ref{sec:prompt_design}.

\noindent \textbf{Max Step.}  We set maximum step limits for agents to evaluate their capabilities of completing tasks within reasonable timeframe.
According to the length of action sequences as shown in Table~\ref{tab:dataset}, we empirically set the maximum step limit as 15 for single-APP and constrained tasks, while for cross-APP tasks, it is set as 30.

\subsection{Poor Performance in Mobile Tasks}
\label{sec:macro_eval}

In this section, we integrate the metrics introduced in \S\ref{sec:macro} to assess the task completion of LLM agents across various task types. 
We report the results across single-APP, cross-APP, and constrained tasks in Table~\ref{tab:macro} and Table~\ref{tab:constrain}. 
Recall that the Success Rate~(SR) is assessed by GPT-4. 
To validate its reliability, we perform cross-validation between it with TR and TCR, where TR and TCR represent alternative perspectives on task completion. 
Specifically, we compute the Pearson Correlation Coefficient~(PCC)~\cite{cohen2009pearson} between SR and TR, resulting in a correlation of $0.87$, and between SR and TCR, yielding a correlation of $0.91$. These high coefficients indicate a substantial correlation between SR and both TR and TCR, \textbf{validating the rationale of adopting the GPT-4 judgment mechanism}.

Table~\ref{tab:macro} reveals a \textbf{significant deficiency of SOTA agents in the real-world mobile tasks}.
While GPT-4 achieves a 75.9\% SR on single-APP tasks, all agents exhibit an inability to make effective decisions across other task settings. 
Noteworthy the performance gap between GPT-4 and GPT-3.5 is much larger for cross-APP tasks than single-APP tasks.
It indicates that the cross-APP tasks are more complex and difficult, and can well reveal the \textbf{significant disparity in planning abilities between the two agents}. 
In contrast to prior benchmark studies, LLaMA2-70B demonstrates inferior performance relative to GPT-3.5 and GPT-4. 
\input{tabs/constrain}

We conduct a detailed examination of the APPs where the SOTA agents, including GPT-3.5 and GPT-4, do not perform well. 
Our investigation reveals \textbf{a vulnerability in handling APPs characterized by deficient textual information and intricate operational logics}. 
To further substantiate this observation, we calculate the PCC between SR and the information richness~(IR) and operation complexity~(OC) of APPs. 
Specifically, we utilize the average length of APP observation and the inverse length of ground truth actions to quantify IR and OC, respectively. 
The results in Table~\ref{tab:app_iroc} demonstrate that OC poses a more substantial challenge for GPT-3.5 in achieving a higher SR, while GPT-4 exhibits a greater sensitivity to IR. 
The high values of IR $\times$ OC further prove our findings.

Beyond basic task completion, we assess the agents' capacity to comprehend constraints and adeptly make decisions to avoid violations. 
Table~\ref{tab:constrain} presents the constraint violation ratios of GPT-3.5 and GPT-4.
LLaMA2 models are excluded as they face challenges in completing basic tasks, rendering this assessment impractical.
Table~\ref{tab:constrain} reveals that even for straightforward constraints, GPT-3.5 still may violate them. 
By reading its intermediate reasoning processes, we discern that GPT-3.5 lacks awareness and understanding of constraints.
For instance, in the case of \textit{``Find the current weather forecast''} with the constraint \textit{``do not use the Weather APP''}, GPT-3.5 directly opens the Weather APP, while GPT-4 comprehends the constraint and devises an alternative way using a web browser to search for weather forecast.
It highlights the \textbf{considerable distance yet to be covered before GPT-3.5 can be applied effectively in permission-sensitive environments}.

\subsection{Four Weakness Leading to Failure}
\label{sec:micro_eval}

\begin{figure}[tbp]
    \centering
    \includegraphics[width=0.76\columnwidth]{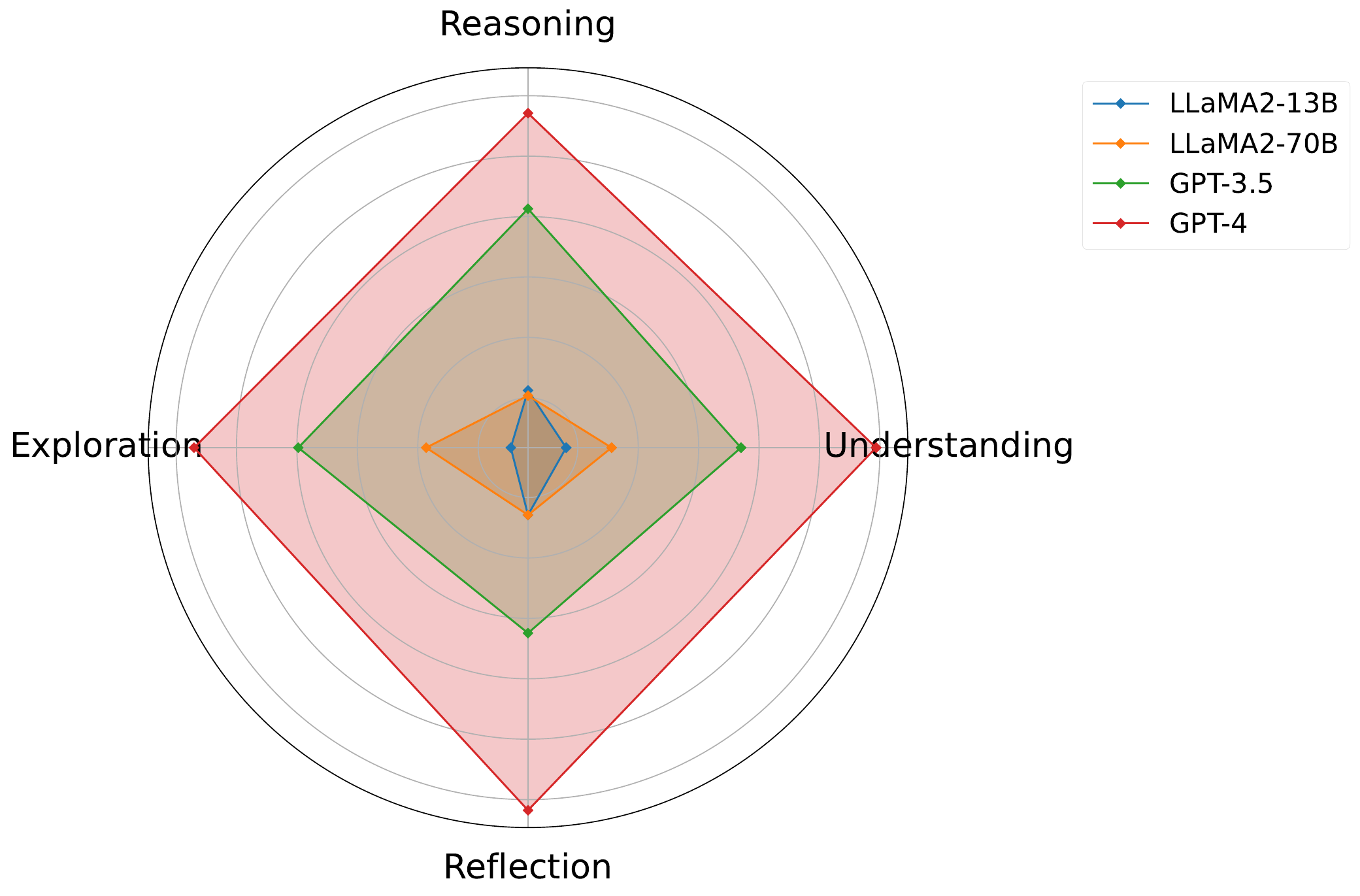}
    \caption{Agent abilities evaluation on cross-APP tasks. }
    \label{fig:micro_eval}
\end{figure}

\begin{figure}[tbp]
	\centering
    	\subfigure[Invalid Action Ratio.]{
    		\includegraphics[width=0.45\columnwidth]{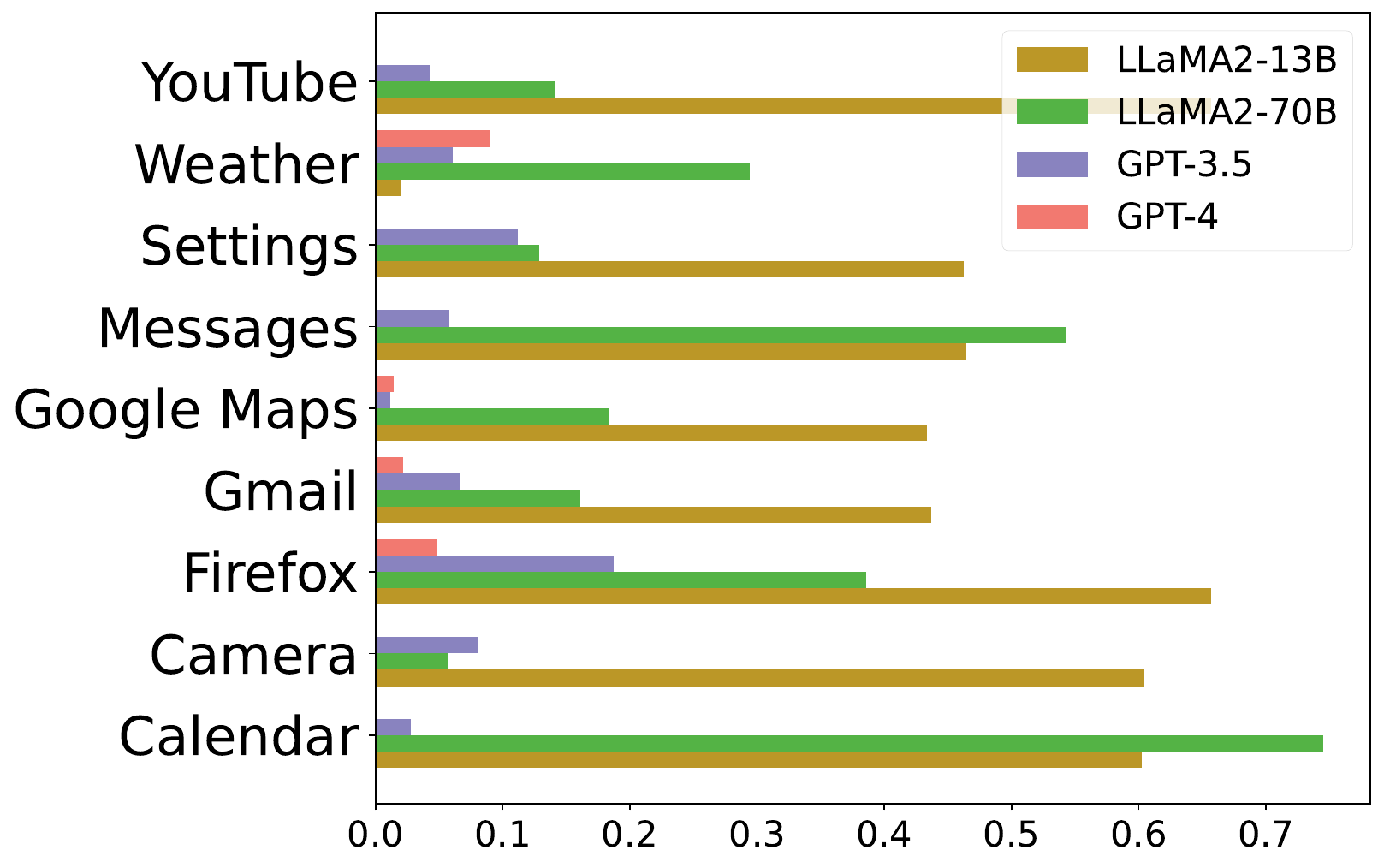}
    		\label{fig:subfig:ia} 
    		}
    	\subfigure[Invalid Format Ratio.]{
    		\includegraphics[width=0.45\columnwidth]{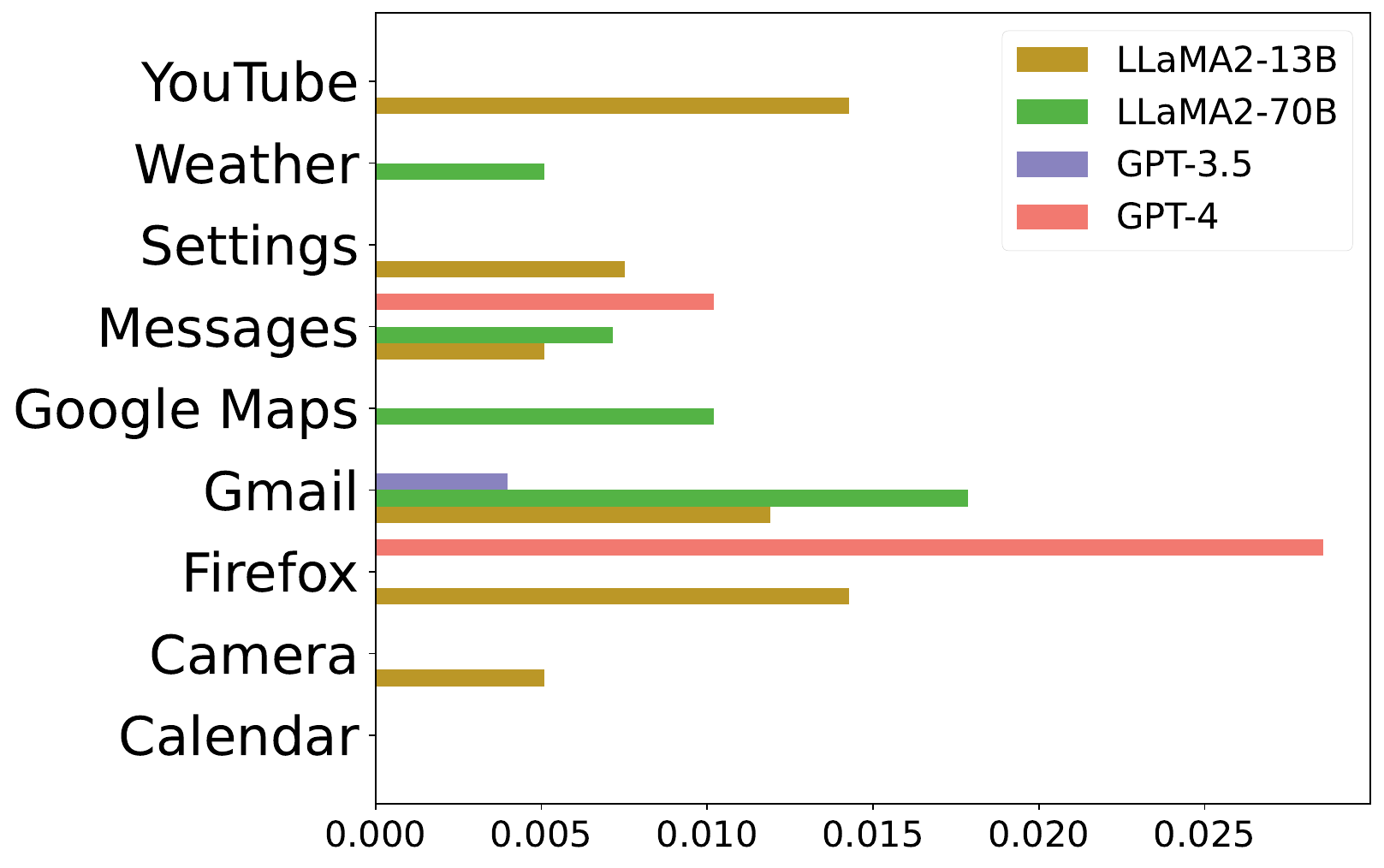}
    		\label{fig:subfig:if} 
    		}

  	\subfigure[Nuggets Mining Score.]{
		\includegraphics[width=0.45\columnwidth]{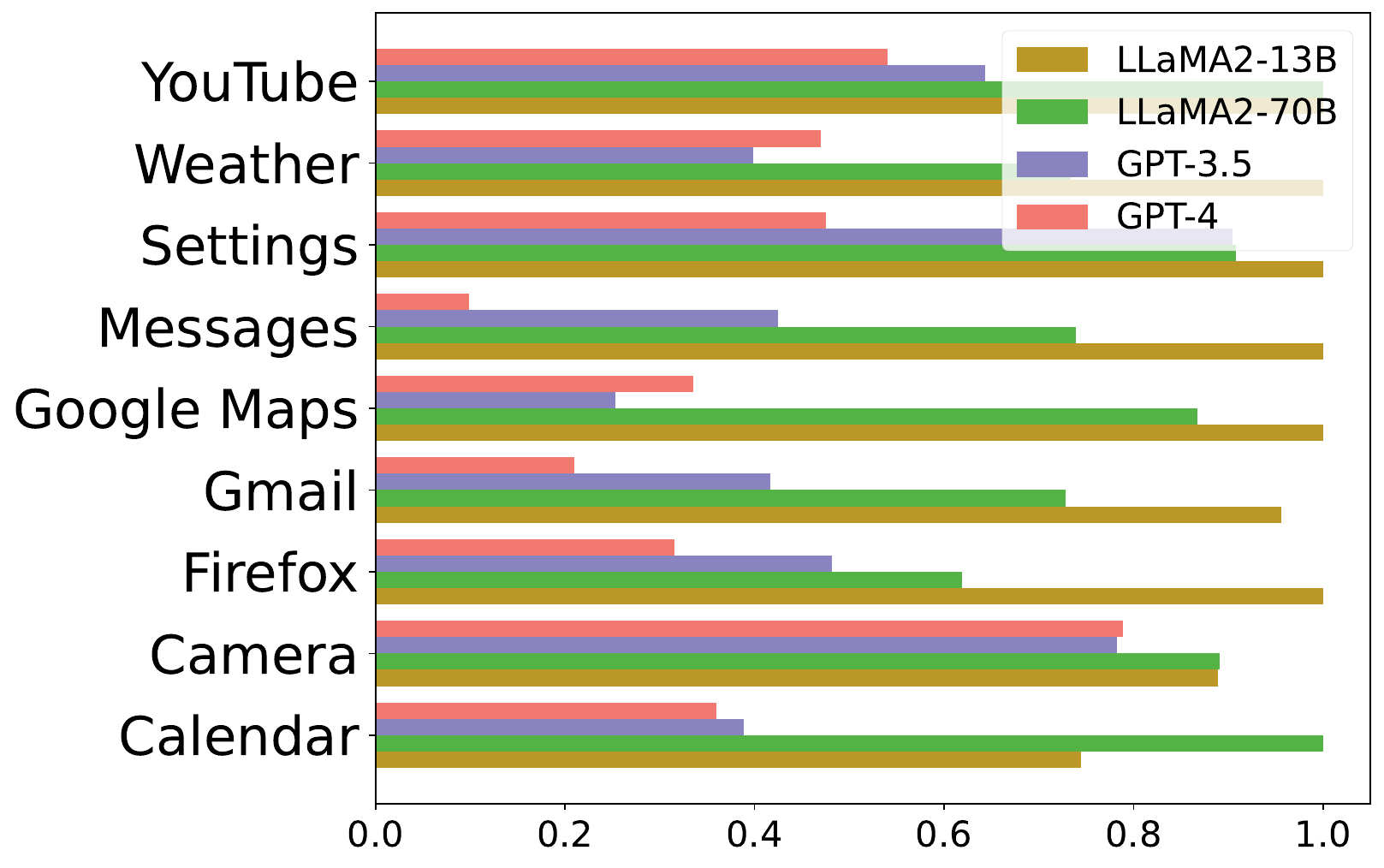}
		\label{fig:subfig:nm} 
		}
	\subfigure[Operation Logic Score.]{
		\includegraphics[width=0.45\columnwidth]{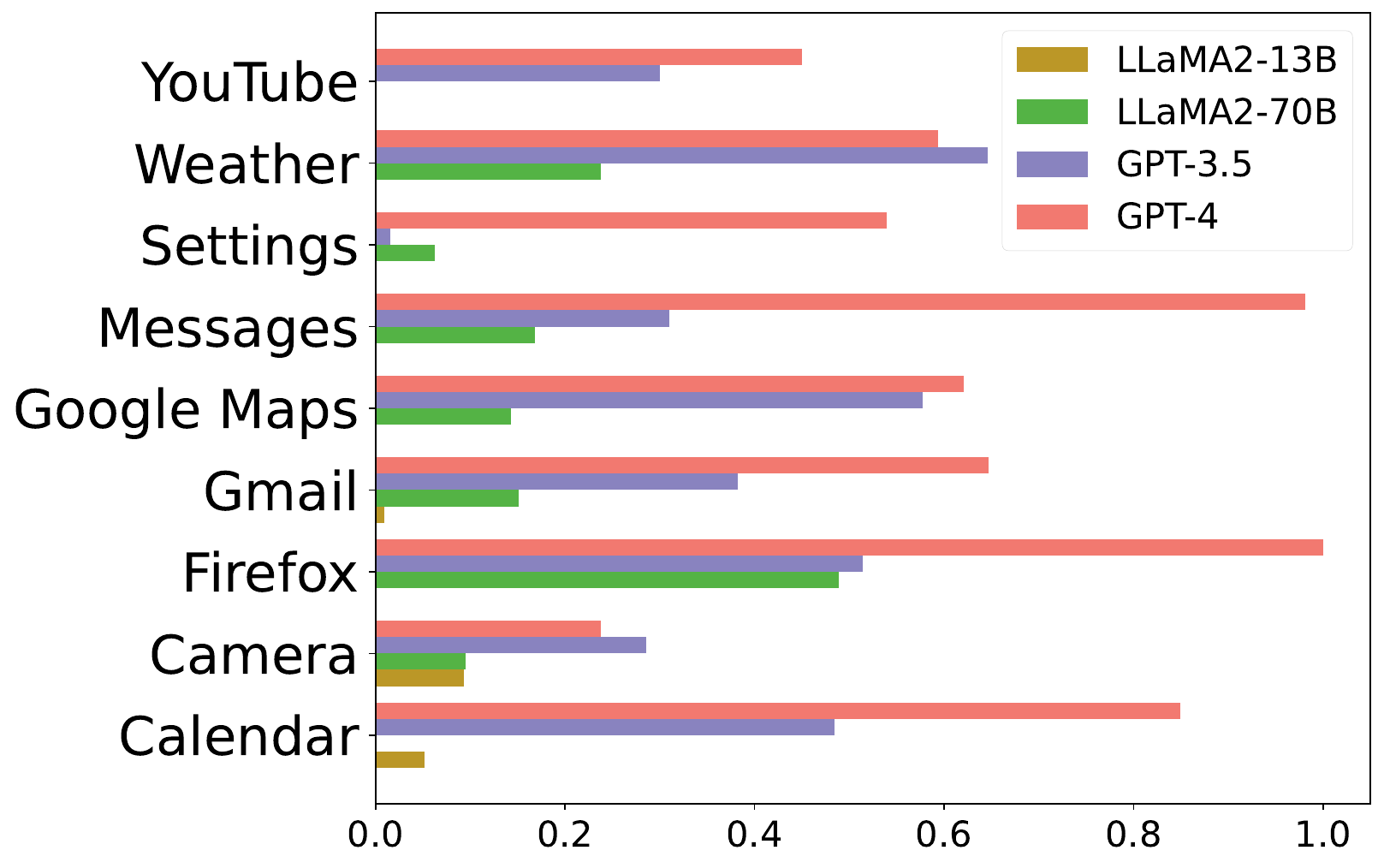}
		\label{fig:subfig:ol} 
		}

  	\subfigure[Aware of Completion Ratio.]{
		\includegraphics[width=0.45\columnwidth]{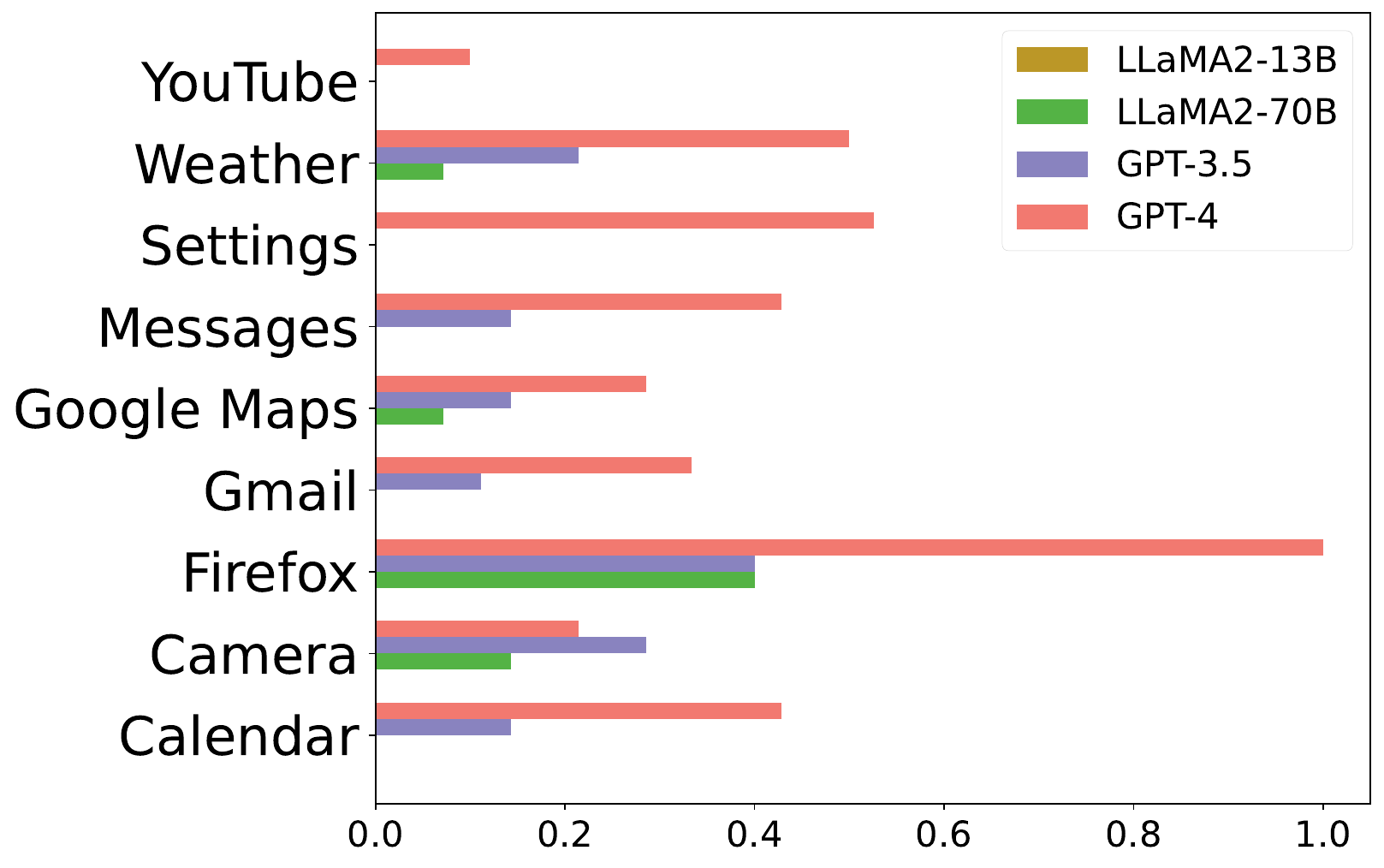}
		\label{fig:subfig:ac} 
		}
	\subfigure[Repeat Action Ratio.]{
		\includegraphics[width=0.45\columnwidth]{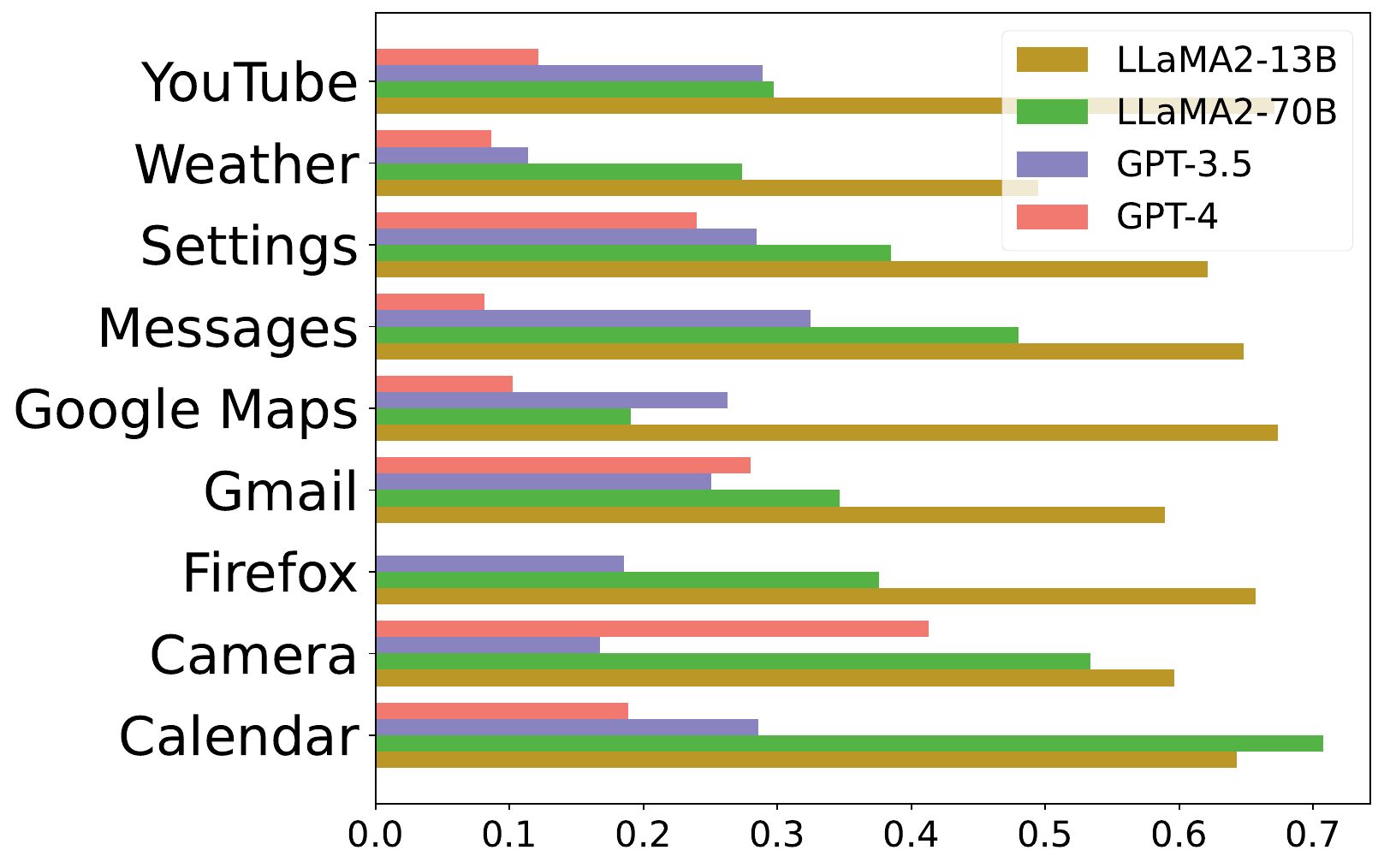}
		\label{fig:subfig:ra} 
		}
	\caption{Metrics for understanding, reasoning and exploration dimensions.}
	\label{fig:micro_metric_eval} 
\end{figure}

In this section, we employ the metrics introduced in \S\ref{sec:micro_metric} to quantify the fine-grained planning abilities of these agents, so as to understand their weaknesses that lead to failure.
As shown in Fig.~\ref{fig:micro_eval}, GPT-4 shows superior performances across various dimensions, further substantiating its excellence in task completion, as indicated in Table~\ref{tab:macro}.
In contrast, \textbf{LLaMA2 models exhibit significant weaknesses across all four dimensions}.
In Fig.~\ref{fig:micro_metric_eval}, we present the composed metrics of these dimensions.
Due to space limit, we only present part of the testing APPs, and the complete APP metrics can be found in \S\ref{sec:micro_app}. 
Fig.~\ref{fig:subfig:ia} and \ref{fig:subfig:if} present the ratios of outputting invalid format and out-of-space actions. 
The notably higher ratios of LLaMA2 show its challenges in understanding and adhering to prescribed action rules.
Fig.~\ref{fig:subfig:nm} demonstrates the superior capacity of GPT-3.5 and GPT-4 to apprehend more nuanced information compared to LLaMA2. 
In Fig.~\ref{fig:subfig:ol}, LLaMA2 agents exhibit challenges in identifying the correct subsequent actions even after multiple attempts. 
Moreover, LLaMA2-13B lacks the awareness that the task has been successfully completed, as depicted in Fig.~\ref{fig:subfig:ac}. 
Fig.~\ref{fig:subfig:ra} indicates a high repeat action ratio of LLaMA2, underscoring its limited ability to explore the environment.
\textbf{GPT-4 also demonstrates a notable proclivity for repeating erroneous actions for several APPs}.
To improve the exploration ability of GPT-4, we introduce an exploration strategy and examine its impact on performance in \S\ref{sec:explore}. 
While GPT-4 shows certain improvement through Reflexion, our analysis suggests that \textbf{it stems from inherent opportunities for additional attempts to complete the task rather than an enhancement in the agent's policy}. Detailed experiments and analysis can be found in \S\ref{sec:reflection}.

\section{Future Directions for Enhancing LLM Agent}
Through experiments in \S\ref{sec:micro_eval}, we observe that LLaMA2 models display weaknesses across all four dimensions. Even for the leading model, GPT-4, still exhibits shortcomings in exploration and reflection. In this section, we first analyze the ineffectiveness of Reflexion and provide an empirical analysis of the factors contributing to this phenomenon. Second, we propose a novel prompt-based exploration method, revealing that explicitly encouraging the agent to explore unknown actions can enhance performance.
\subsection{Analysis of Reflection's Failure}
\label{sec:reflection}
\begin{figure}[tbp]
	\centering
	\subfigure[SR of GPT-3.5 agent.]{
		\includegraphics[width=0.45\columnwidth]{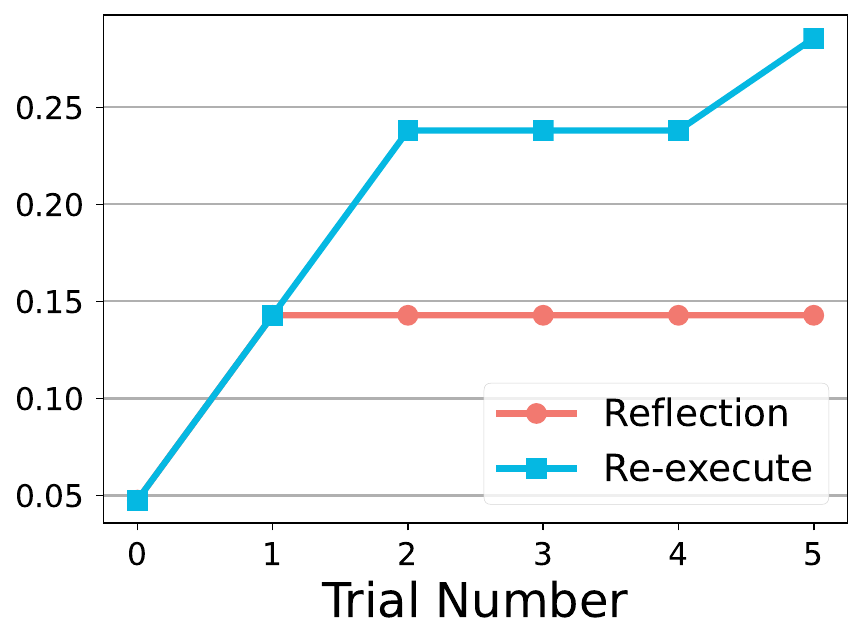}
		\label{fig:subfig:reflect_tr} 
		}
	\subfigure[SR of GPT-4 agent.]{
		\includegraphics[width=0.45\columnwidth]{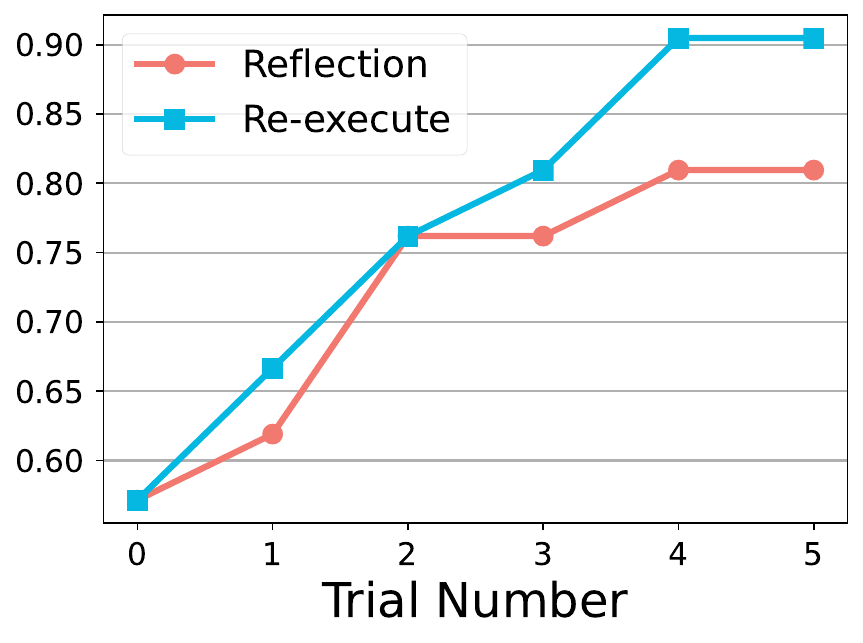}
		\label{fig:subfig:reflect_tcr} 
		}
	\caption{Performances evaluated on cross-APP tasks by increasing the reflection times.
	}
	\label{fig:reflect_eval} 
\end{figure}

Recall that Reflexion summarizes experience and then re-executes failed tasks, it inherently offers opportunities for additional attempts and possesses potential for performance improvement. 
Accordingly, we conduct a comparative evaluation with re-executing failed tasks without the Reflexion process, namely \textit{Re-execute}.
In Fig.~\ref{fig:reflect_eval}, we present the SR of Reflexion@5 and Re-execute@5. 
Contrary to expectations, we observe that \textbf{Reflexion does not yield positive outcomes compared to Re-execute}. 
This unexpected phenomenon motivates an investigation of the underlying mechanisms of Reflexion and the challenges of applying it in our scenario.
To initiate our investigation, we provide a formal definition of the Reflexion process, specified as follows:
\vspace{-10pt}
\begin{equation}
    \underbrace{P(\text{new\ trajectory}\ |\ \text{reflection})}_{\text{\ding{174}}}\cdot \underbrace{P(\text{reflection}\ |\ \overbrace{\text{old\ trajectory}}^{\text{\ding{172}}})}_{\text{\ding{173}}}, 
    \label{eq:reflect_process}\nonumber
\vspace{-5pt}
\end{equation}
where $P$ denotes the LLM agents.
This equation describes the Reflexion process, \ie extracting valuable insights from past trajectories and benefiting subsequent trials. 
Three key steps in this equation may contribute to the degradation of Reflexion performance in our scenario. 
The first and \textbf{the most important reason is that the old trajectory~(marked in \ding{172}) is less informative} compared with previous scenarios.
Unlike benchmarks~\cite{austin2021program,yang2018hotpotqa} characterized by one-step decision, and virtual ALFWorld environment~\cite{shridhar2020alfworld} with small and static action space, our environment necessitates multi-step planning within a vast and dynamic action space. 
This challenge not only makes it hard to explore the entire action space, thus cannot provide sufficient guidance for next trial, but also sparsifies the reward feedback.
Therefore, a potential improvement can be achieved by improving trace quality. 
In specific, employing exploration strategies to broaden the explored action space for informative experience~\cite{ma2023laser} and devising intrinsic rewards to mitigate the sparse reward issue~\cite{yuan2023plan4mc}.
Secondly, a constrained ability to distill reflection (\ie part \ding{173}) diminishes the reflection efficacy. 
Lastly, regarding part \ding{174}, the reflection may not be fully leveraged by agent or, conversely, introduces bias~\cite{huang2023large} and degrades the performance compared to the Re-execute that is without reflection.

\subsection{Enhancing Exploration Boosts Performance}
\label{sec:explore}
\begin{figure}[tbp]
	\centering
	\subfigure[Success rate.]{
		\includegraphics[width=0.45\columnwidth]{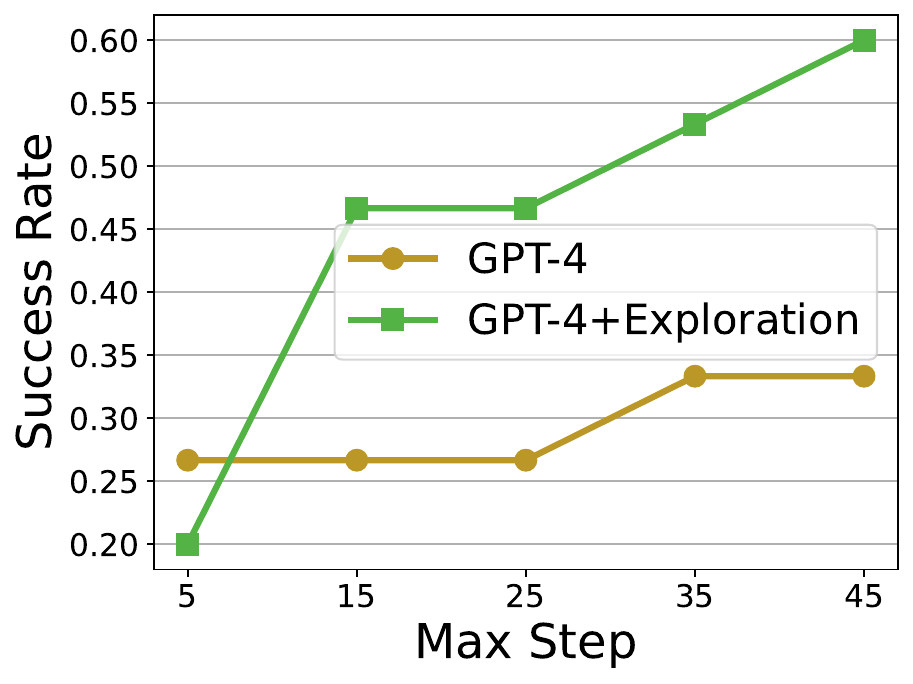}
		\label{fig:subfig:explore_tr} 
		}
	\subfigure[Repeat action ratio.]{
		\includegraphics[width=0.45\columnwidth]{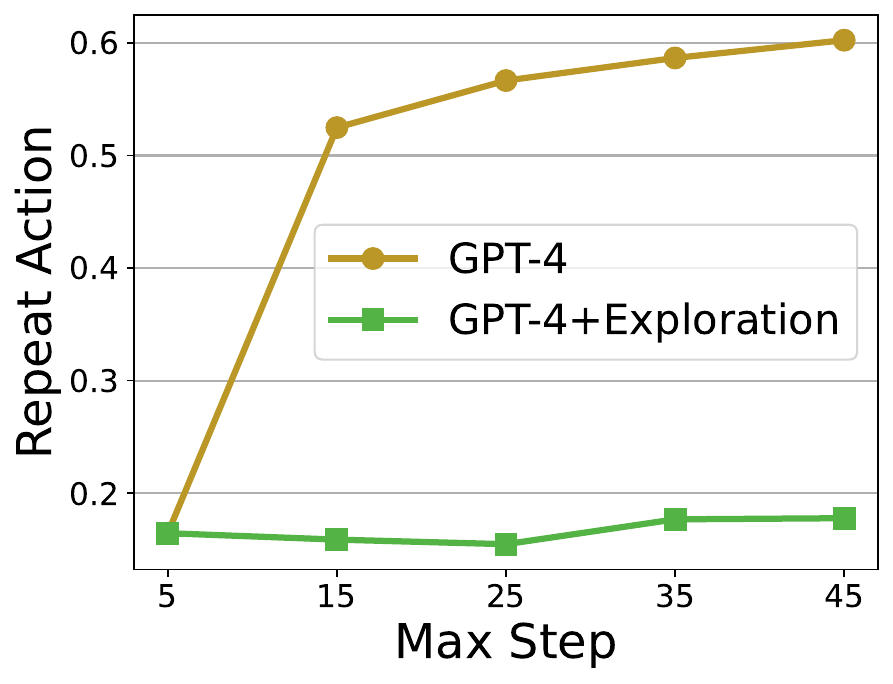}
		\label{fig:subfig:explore_explore} 
		}
	\caption{GPT-4 and GPT-4+Exploration comparisons by varying the maximum step limit on the Camera APP.
	}
	\label{fig:explore_eval} 
 \vspace{-10pt}
\end{figure}

Upon reading the trajectories, we observe that even for GPT-4, it still presents a pronounced tendency to repeat erroneous actions, as illustrated in Fig.~\ref{fig:subfig:ra}, indicating its limited exploration capabilities. 
Furthermore, the repetition of actions degrades the quality of preceding trajectories, rendering them insufficient for providing informative guidance for reflection, as discussed in \S\ref{sec:reflection}.

In this section, we introduce a novel prompt-based exploration strategy for LLM agents. 
Diverging from prior approaches~\cite{carta2023grounding} that treat the LLM as a RL policy network and employ exploration strategies originating from RL, our strategy guides the exploration of LLM agent by \textbf{incorporating a prompt indicating the count of previously visited observations $M(\text{{s}})$ and issued actions $N(\text{{s, a}})$}.
Specifically, we embed a hint prompt such as \textit{``You have already been in the current state M times, and taken action A for N times''} at each decision step. 
This concept is inspired by the Upper Confidence Bound~(UCB)~\cite{auer2002finite, improving_ucb}. Unlike UCB, we do not design explicit exploration strategies. Instead, we integrate historical information into the prompt, leveraging the powerful decision-making capabilities of the LLM to balance exploration and exploitation.
We conduct an experiment on the Camera APP, where GPT-4 exhibits the highest repeat action ratio, to evaluate the effectiveness of the exploration strategy. 
We vary the maximum step limit in $\{5, 15, 25, 35, 45\}$, and present SR and repeat action ratio in Fig.~\ref{fig:explore_eval}.
The results show that, with a simple counting-based prompt, SR can achieve 27\% of improvement. 
Furthermore, as the maximum step limit increases, the exploration ability of GPT-4 degrades. In contrast, for GPT-4+Exploration, effective exploration of the environment persists, leading to continued performance improvement.

%% file: tabs/constrain.tex
\begin{table}[tbp]
\centering
\caption{Constraints violation ratios for Constrained Tasks.}
\label{tab:constrain}
\begin{tabular}{cccc}
\hline
Model   & APP-level & Page-level & Component-level \\ \hline
GPT-3.5 & 0.207                                     & 0.072     & 0.33                                 \\
GPT-4   & 0.000                                     & 0.050       &0.00                               \\ \hline
\end{tabular}
\end{table}

%% file: sections/conclusion.tex
\section{Conclusion}
This study introduces \ours environment and a scalable benchmark.
It supports the evaluation of cross-APP and constrained task scenarios. 
We propose adaptive and precise metrics to assess task completion, and fine-grained abilities of agents to understand their weaknesses. 
The results underscore significant room for improvement among SOTA agents. 
We highlight four research directions for enhancing LLM agents. 
Additionally, we offer empirical insights into the failure of reflection and present a novel method to enhance the exploration capabilities of agents. 
In the future, we plan to investigate the weaknesses of multi-modal model agents. 
Given that vision models excel at spatial understanding and reasoning, areas where LLMs struggle~\cite{yamada2023evaluating}, we intend to scrutinize their fine-grained abilities and identify promising research directions in this domain. 
Our \ours supports multi-modal evaluation, and the benchmark can be easily extended to this setting.

%% file: sections/appendix.tex
\section*{Appendix}

\section{Detailed Observation Compression Method}
\label{sec:detailed_compress_method}

\begin{figure*}[htbp]
	\centering
	\subfigure[Screenshot.]{
		\includegraphics[width=0.16\textwidth]{figs/ob_1.pdf}
		\label{fig:subfig:obs1} 
		}
	\subfigure[XML derived by UIAutomator.]{
		\includegraphics[width=0.41\textwidth]{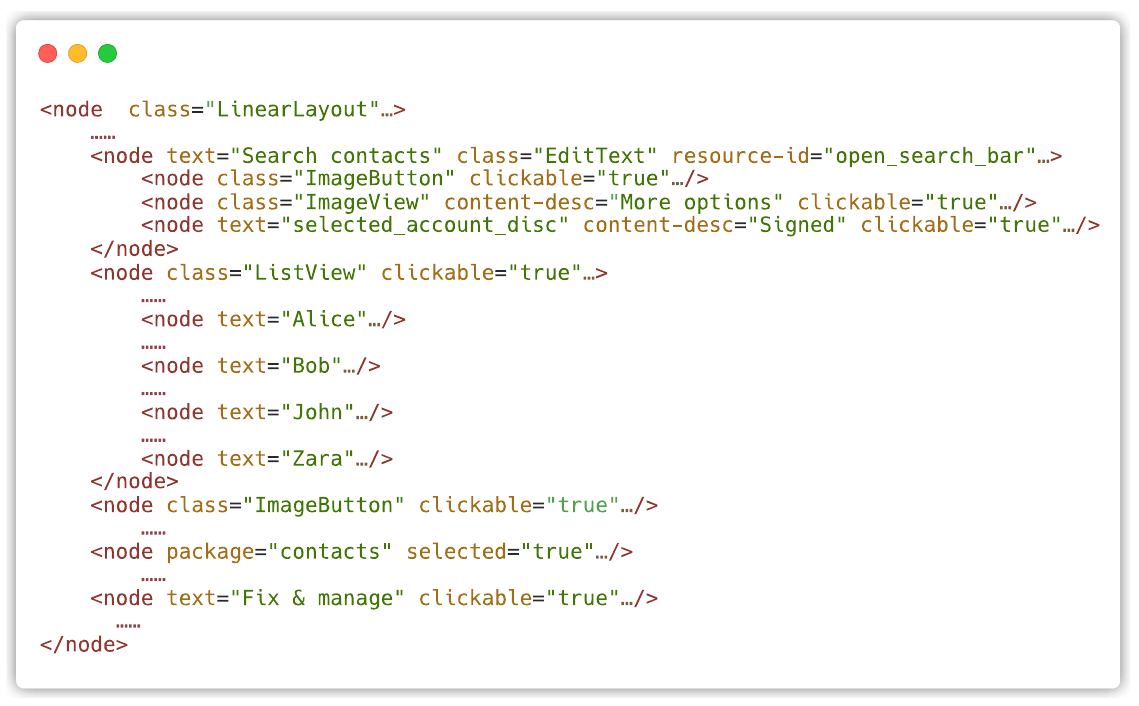}
		\label{fig:subfig:obs2} 
		}
 	\subfigure[Compressed observation.]{
		\includegraphics[width=0.39\textwidth]{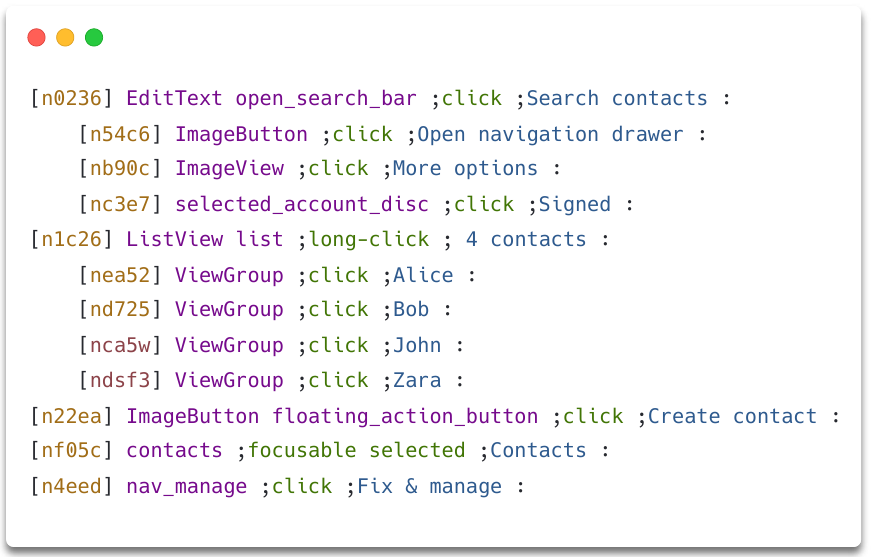}
		\label{fig:subfig:obs3} 
		}
	\caption{An example of the screenshot, original XML and compressed observation of Contacts APP page.
	}
	\label{fig:xml_example} 
\end{figure*}

The textual observation is derived from the XML representation encapsulating comprehensive screen information as shown in Fig.~\ref{fig:xml_example}. However, directly inputting the entire XML into the LLM, proves to be excessively lengthy as illustrated in Table~\ref{tab:obs_len}. To mitigate this, we employ a two-phase heuristic approach for compressing the XML to a manageable length for LLM processing. The XML entries are categorized into two groups: one for layout, which does not support actionable operations, and the other for UI components. We eliminate the XML entries related to layout, retaining only those associated with UI components. In the second phase, we merge non-functional and non-visible nodes upwards, incorporating their descriptive information into the parent nodes. This strategy enhances the LLM's ability to understand the semantic of the hierarchical XML tree, and result in a more efficient compression. 
For components with nuanced state descriptions, we amplify their textual information. For instance, when the switch component is in the off position, we append a description stating \textit{``it is currently unchecked, and you can switch it on."}.
To enable the agent to accurately select the UI component for operation, a unique ID is assigned to each component in the compressed observation. In the compressed observation, components are structurally organized, maintaining their ancestral-descendant relationships in the original XML tree, aiding the LLM agent in comprehending the interface's layout through text and thereby enhancing its command efficacy.

\begin{table}[htbp]
\centering
\caption{We randomly select several APPs and compare the token numbers before and after compression. Our compression ratio reach a high of 86.6\%, while preserving the semantic information. This approach enhances the utilization of the LLM agent context, allowing for the accommodation of more historical observations in each decision-making process.}
\resizebox{\columnwidth}{!}{%
\begin{tabular}{ccc}
\hline
\textbf{App Name} &  \textbf{\#Token (Original)} &  \textbf{\#Token (Compressed)} \\
\hline
Gmail (email\ list) & 11,707 & 1,155 \\
Gmail (compose\ email)  & 7,273 & 413 \\
Calendar & 8,604 & 584 \\
Google map & 15,725 & 637 \\
YouTube & 12,005 & 939 \\
Play Store & 10,450 & 620 \\
Google drive & 11,060 & 651 \\
Clock Alarm & 9,633 & 505 \\
Clock & 7,980 & 285 \\
\hline
\end{tabular}
}
\label{tab:obs_len}
\end{table}

\section{Detailed Action Space}
\label{sec:Detailed_action_space}

We support four-level's action space, \ie APP level, component level, system level and task level. App level actions are responsible for installing, launching and stopping APPs. Most actions are component level which are responsible for operating UI components, such as clicking, typing, and swiping. We also support system level actions including turning the screen on and off, adjusting the volume, setting orientation, and taking screenshots.
Task-level action is designed for the agent  to decide if a task should finish. 

In previous work like AndroidEnv, the action is done by successive touches and lifts, each consists a position $(x,y)$ and an $ActionType \in \{TOUCH, LIFT, REPEAT\}$. AndroidEnv divides the screen into a grid and restricts the ActionType to TOUCH, or groups action sequences like [TOUCH, LIFT, TOUCH, LIFT] into a single gesture, such as swiping, scrolling, or drag-and-drop. However, continuous touches and lifts bring additional inference overhead for agents, and cannot accurately simulate the continuity and smoothness of swiping. 
Instead of interacting with the phone by successive touches and lifts, we directly operate the UI components of APPs through UIAutomator.
It is a testing framework for Android, sending a series of events including pressing, dragging, and scrolling. These events are consistent with real finger slides. Operating components by sending action events is not only more accurate and natural in simulating real user operations, but also superior in terms of APP compatibility.
We can get the executable actions that each component can perform from the corresponding XML and maintain them in the compressed observation, such as clickable, double clickable, long clickable, etc. At the same time, we also record the type of each component in the compressed observation, such as button, text-editor, which can assist the agent to give appropriate action instructions. As we have set a unique ID for each component in the compressed observation, the agent can operate a component by specifying its ID and the corresponding action type.
Since UIAutomator locates and operates on components based on their XPath, our implementation employs a mapping table to convert component IDs into component XPaths, after which we perform the operations.

\begin{table}[tbp]
\centering
\caption{The complete action space, including action type and the corresponding parameters.}
\resizebox{\columnwidth}{!}{%
\begin{tabular}{ccc}
\hline
\textbf{Action level} & \small \textbf{Action Type} & \small \textbf{Action Parameters} \\ \hline
\multirow{4}{*}{APP level} & Install APP & Download link \\
 & Launch APP & Package name \\
 & Stop APP & Package name \\
 & Stop all APP &  \\ \hline
\multirow{7}{*}{Component level} & Click & XPath \\
& Double click  & XPath \\
& Long click  & XPath \\
& Set text  & XPath,\ Text \\
& Swipe up/down/left/right  & Number \\
& Press back  & \\ 
& Press home  & \\ \hline
\multirow{4}{*}{System level}    &
Screen on/off   &
\\ 
&
Volume up/down/mute   &
\\ 
&
Set orientation    &
Horizontal/vertical 
\\ 
&
Screenshot    &
\\ \hline
Task Level     &
Finish task    &
\\ \hline
\end{tabular}
}
\label{tab:action_space}
\end{table}

\section{Benchmark Statistics}
\label{sec:benchmark_stat}
\begin{figure}[tbp]
\centering
\includegraphics[width=0.7\columnwidth]{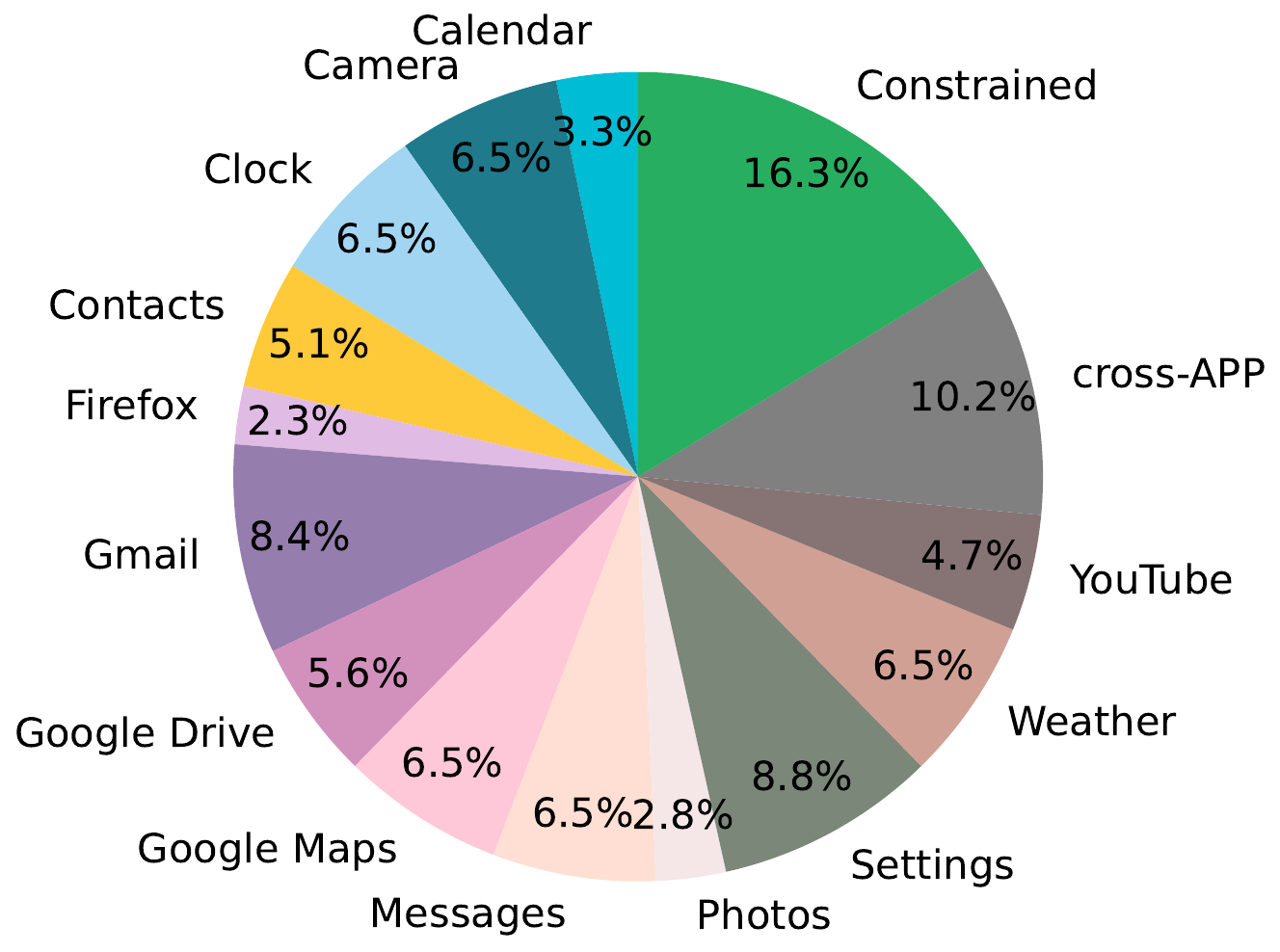}
\caption{Proportion of instructions for different Apps}
\label{fig:detailed_benchmark}
\end{figure}

Our benchmark comprises three types of tasks: single-APP, cross-APP, and constrained tasks. 
The single-APP tasks are derived from 13 APPs including Calendar, Camera, Clock, Contacts, YouTube, Weather, Settings, Photos, Messages, Google Maps, Google Drive, Gmail and Firefox. 
Statistical information regarding the benchmark is illustrated in Fig.~\ref{fig:detailed_benchmark}.

\section{Planning ability scores}
\label{sec:micro_compute}
In this section, we present a detailed explanation of the calculation method for the four fine-grained planning ability scores introduced in \S\ref{sec:micro_metric}. 
The value ranges of these metrics are $[0,\ 1]$, with the exception of Nuggets Mining, Operation Logic, and Reflexion@K. 
Following AgentBench~\cite{liu2023agentbench}, we normalize the latter three metrics to $[0,\ 1]$ across all models. 
It is important to note that smaller metrics in understanding and exploration indicate better performance in these dimensions, while larger values for the metrics in reasoning and reflection denote superior performance in those aspects. 
The calculations for the four dimensions are specified as follow:
\begin{align}
Understanding &= (1-Invalid\ Format\ Ratio) + \nonumber \\&(1-Invalid\ Action\ Ratio) + (1-  Nuggets\ Mining) \nonumber \\
Reasoning&=Operation\ Logic + Awareness\ of\ Completion \nonumber\\
Exploration&=1-Repeat\ Action\ Ratio \nonumber \\
Reflection&=Reflexion@K \nonumber
\end{align} 
Subsequently, we standardize the four dimension scores and present them in Fig.~\ref{fig:micro_eval}.

\section{Complete Results for Testing APPs}
\begin{figure}[tbp]
	\centering
    	\subfigure[Invalid Action Ratio.]{
    		\includegraphics[width=0.45\columnwidth]{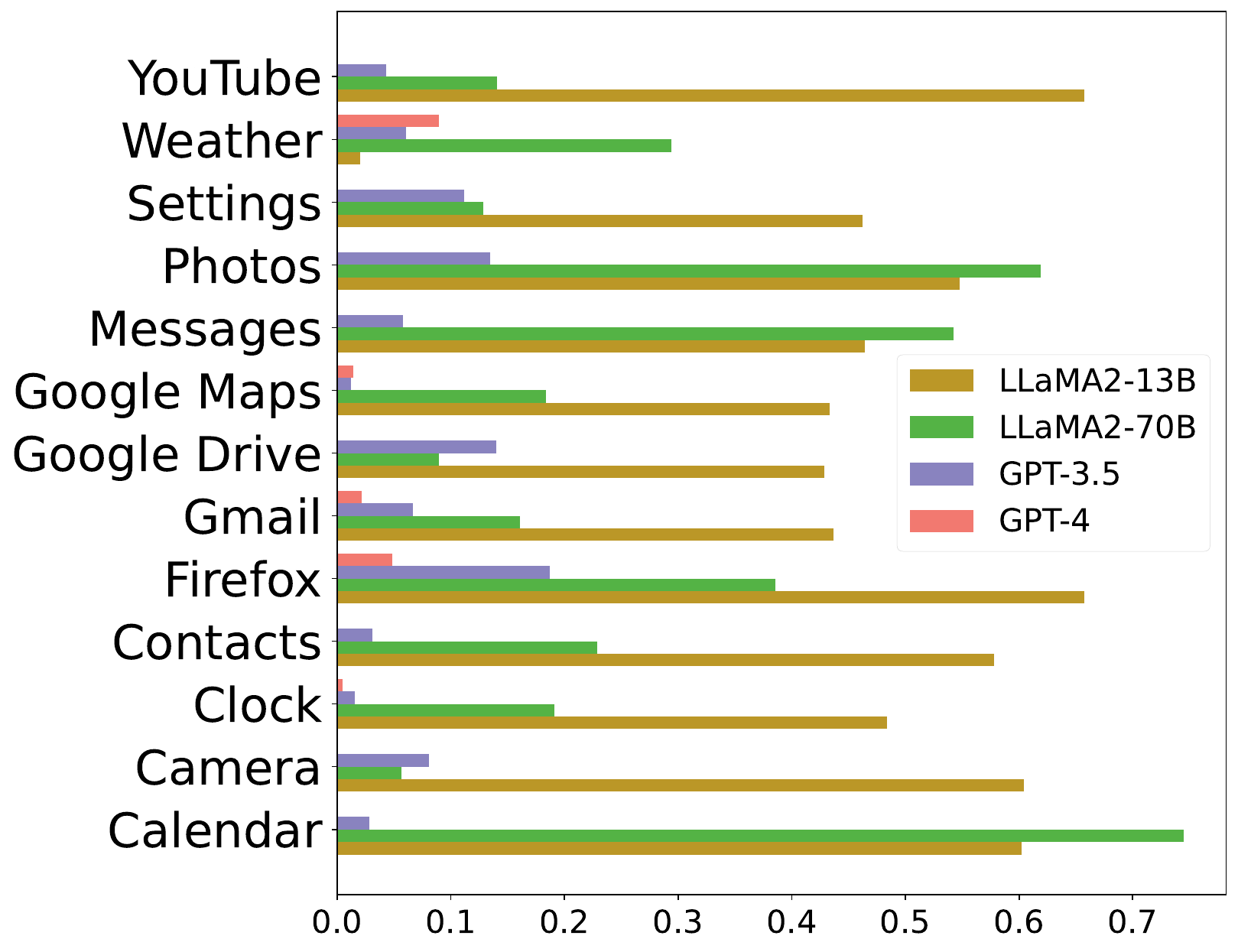}
    		}
    	\subfigure[Invalid Format Ratio.]{
    		\includegraphics[width=0.45\columnwidth]{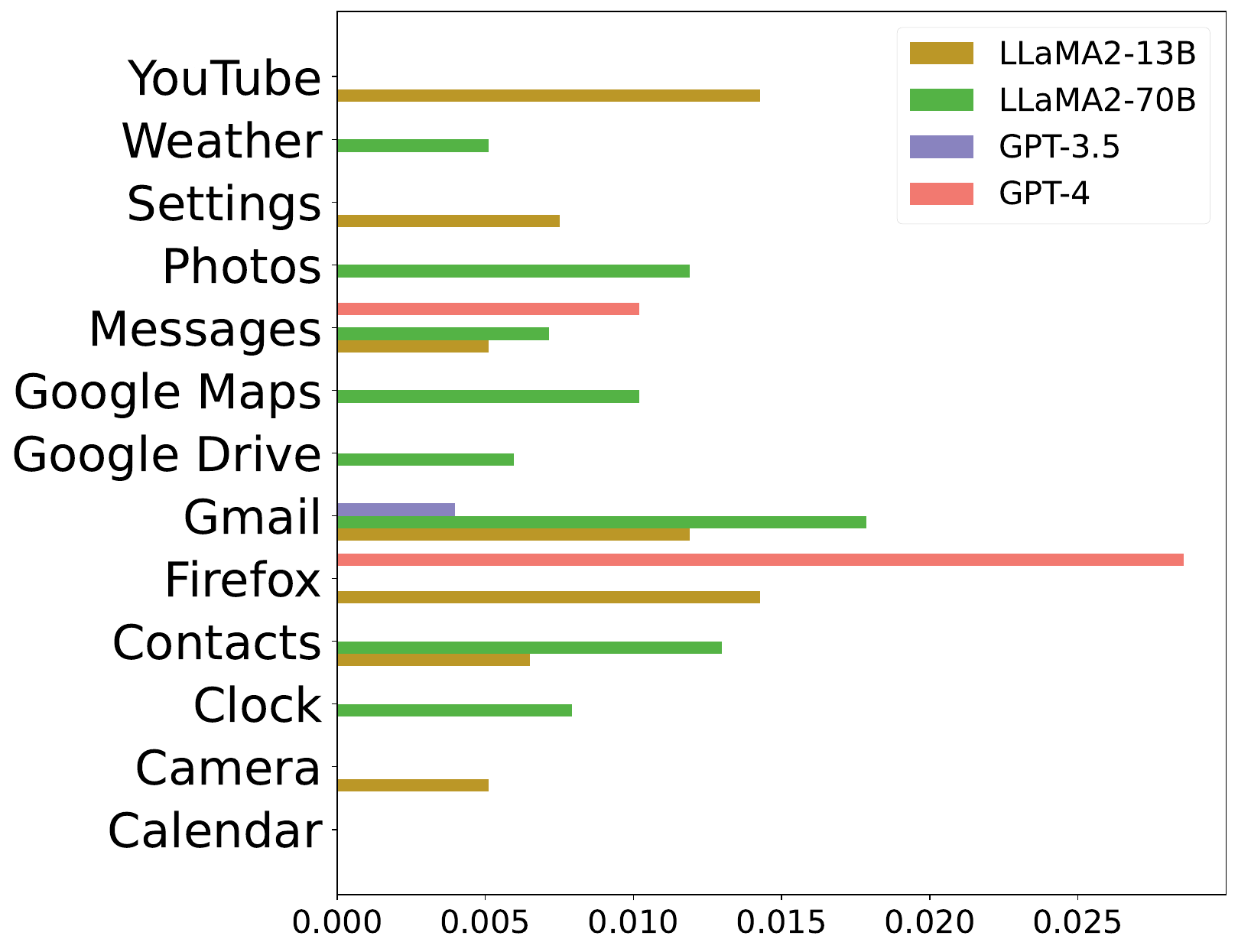}
    		}

  	\subfigure[Nuggets Mining Score.]{
		\includegraphics[width=0.45\columnwidth]{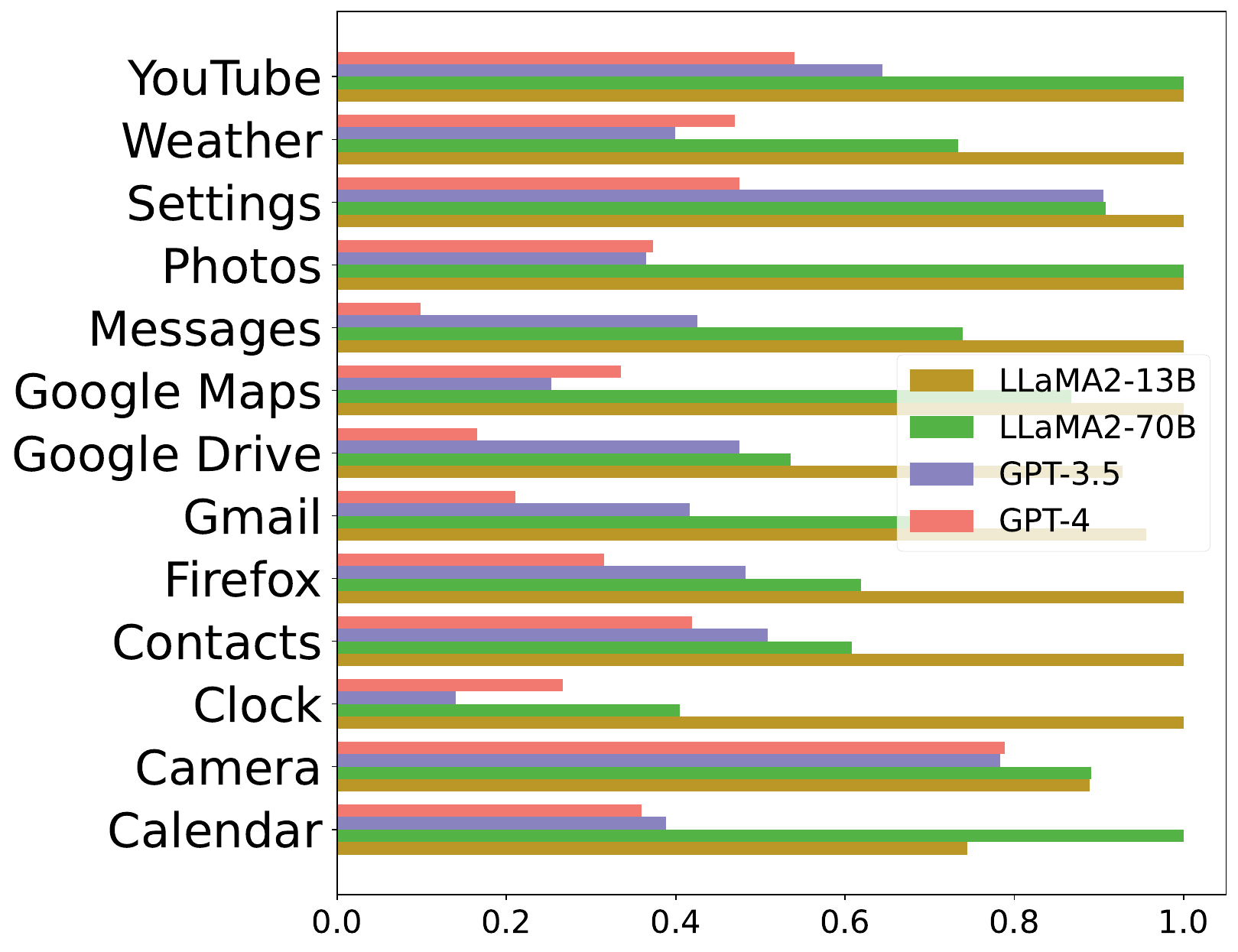}
		}
	\subfigure[Operation Logic Score.]{
		\includegraphics[width=0.45\columnwidth]{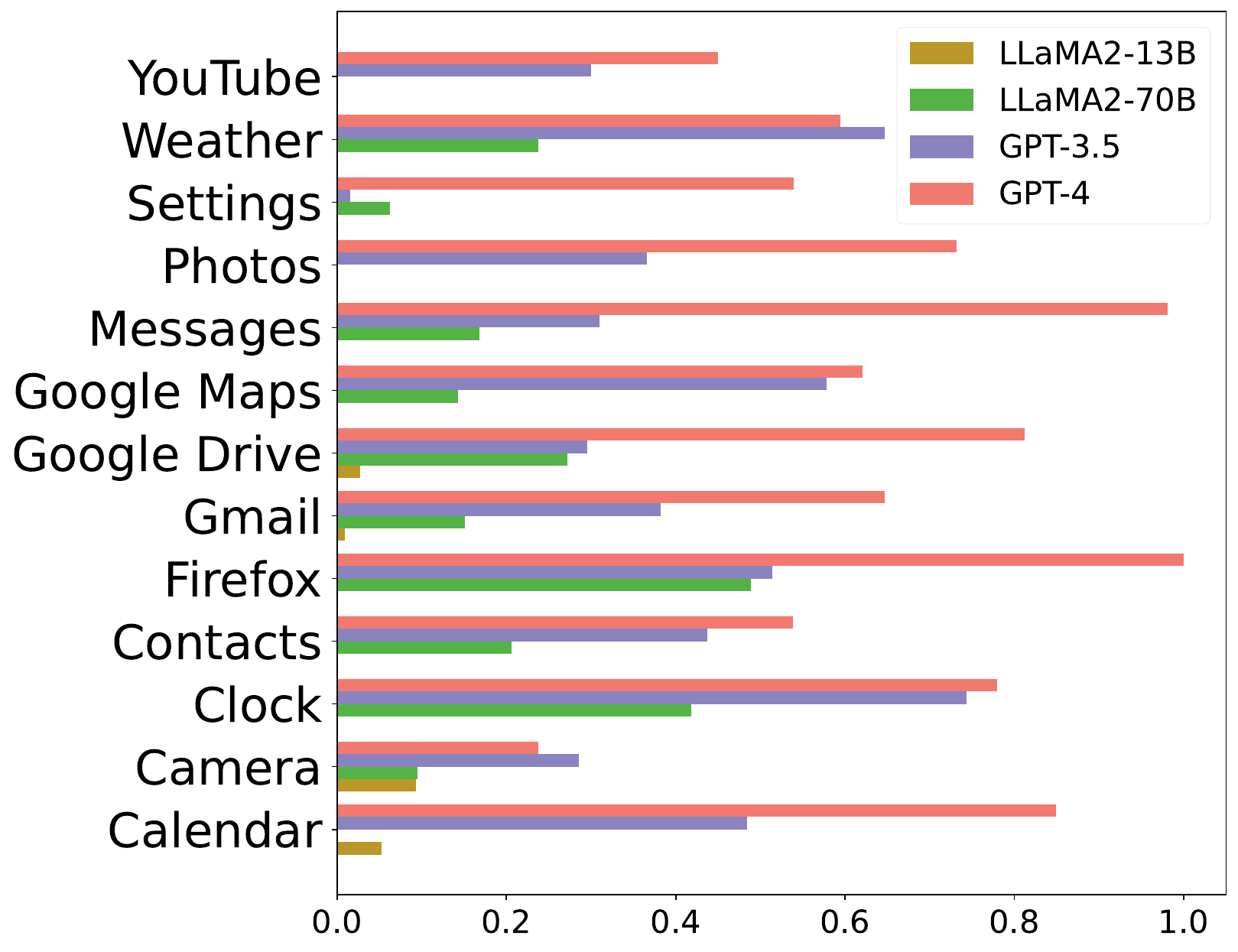}
		}

  	\subfigure[Aware of Completion Ratio.]{
		\includegraphics[width=0.45\columnwidth]{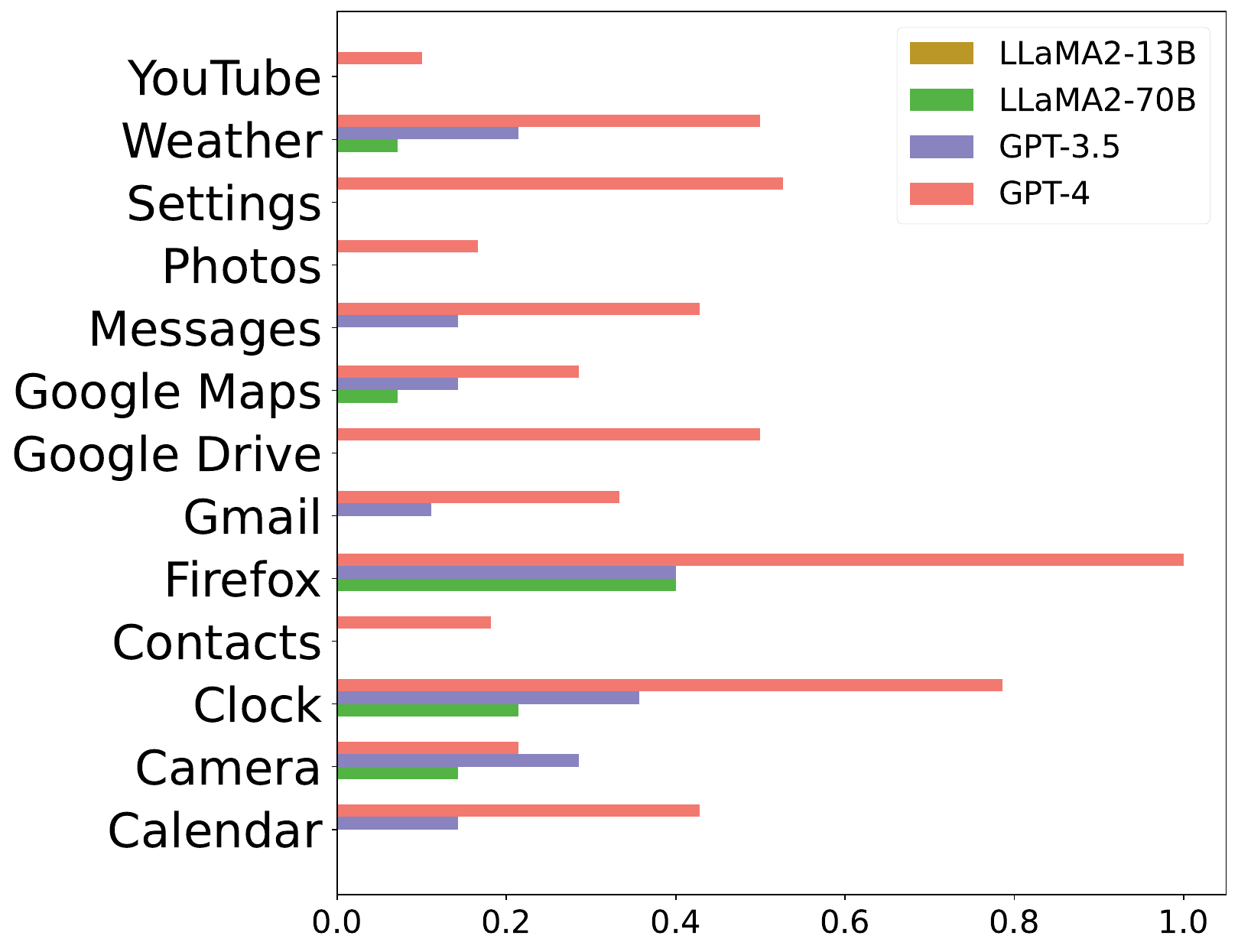}
		}
	\subfigure[Repeat Action Ratio.]{
		\includegraphics[width=0.45\columnwidth]{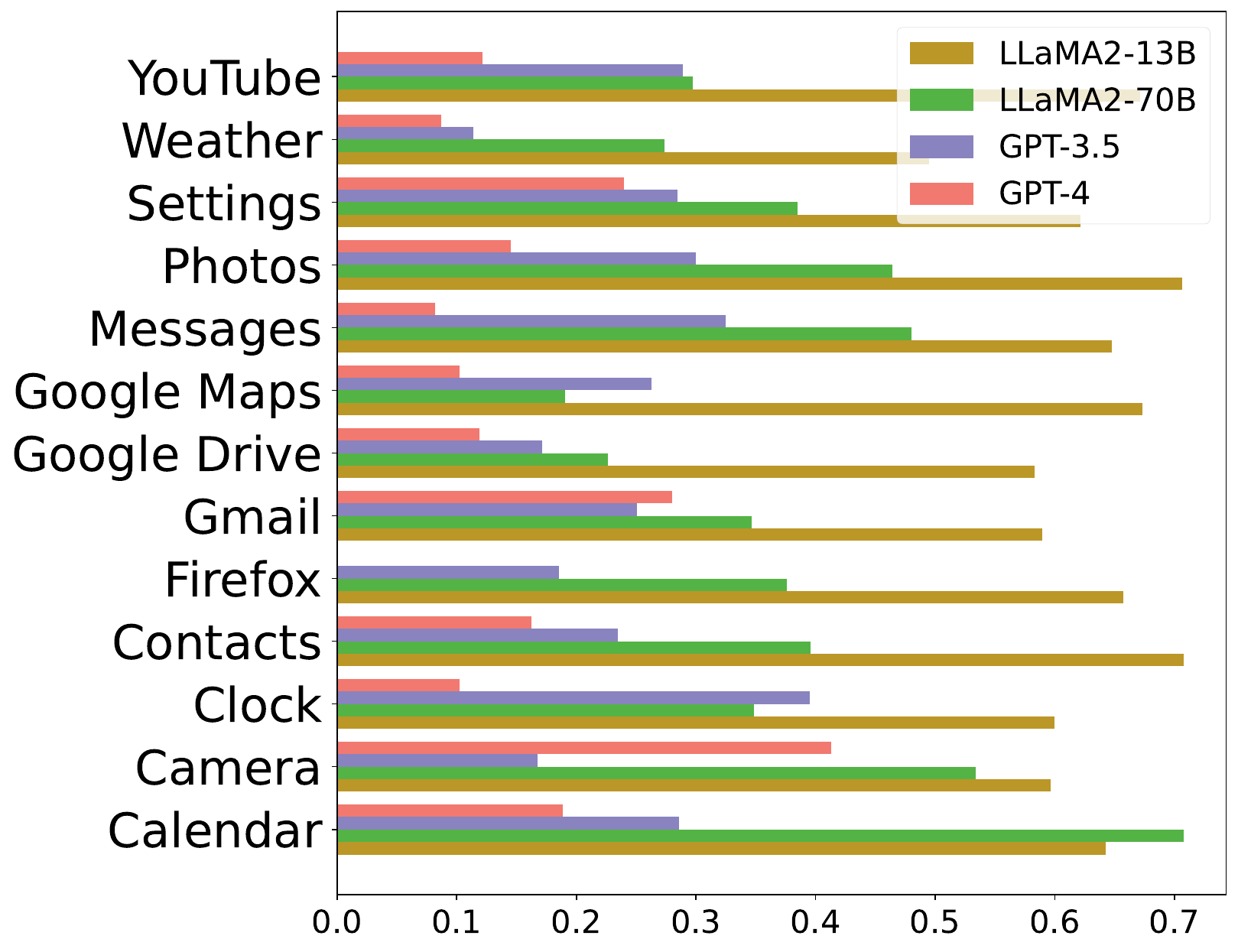}
		}
	\caption{Metrics for understanding, reasoning and exploration dimensions on all testing APPs.}
	\label{fig:micro_metric_eval_all} 
\end{figure}

Due to space limit, we present part of the testing APPs in \S\ref{sec:macro_eval}.
In Fig.~\ref{fig:micro_metric_eval_all}, we show the metrics for all testing APPs.
\label{sec:micro_app}

\section{Prompt Design}
\label{sec:prompt_design}
\begin{figure}[tbp]
\centering
\includegraphics[width=\columnwidth]{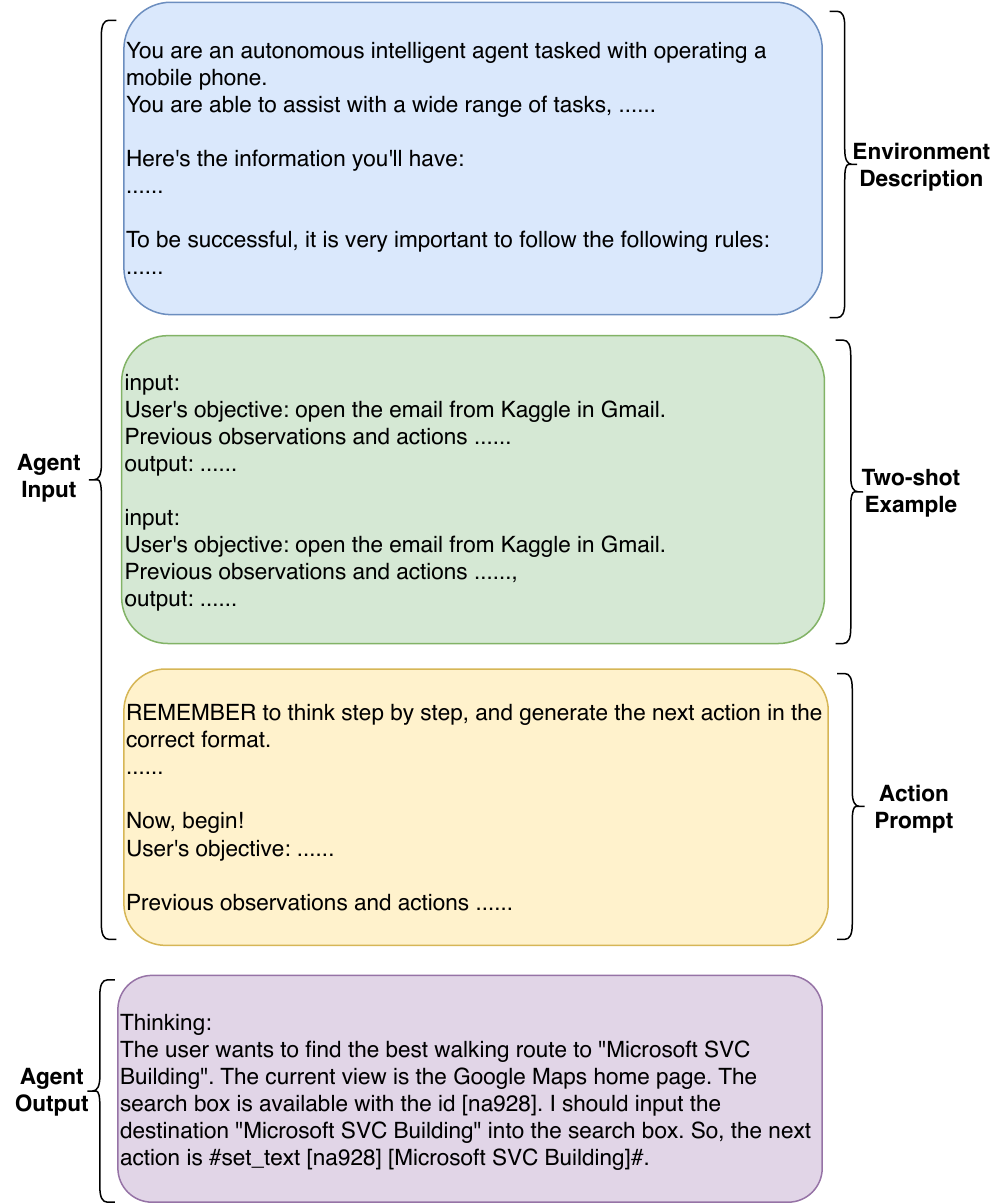}
\caption{Prompt structure.}
\label{fig:prompt}
\end{figure}

Following WebArena, our prompt for each decision-making step incorporates environment descriptions, two-shot examples, task instructions, and historical observations and actions, as illustrated in Fig.~\ref{fig:prompt}. 
We adopt the Reflexion prompt from its official implementation with modifications tailored to our specific scenario. 
The prompt context limit is 4K for LLaMA2-13B, LLaMA2-70B, and GPT-3.5, and at 8K for GPT-4. 
Given that historical observations and actions may exceed the context limit, and to ensure a fair comparison, we apply the truncation strategy employed in WebArena across all agents to maintain a prompt within the 4K context limit.

\subsection{Environment Description Prompt}
\label{sec:env_prompt}
\begin{tcolorbox}[title=Environment Description Prompt, breakable]
You are an autonomous intelligent agent tasked with operating a mobile phone. 
You are able to assist with a wide range of tasks, from answering simple questions to planning and executing a complicated instruction with specific actions you can issue. 

Here's the information you'll have:\\
The user's objective: This is the task you're trying to complete.\\
The installed APPs: These are the APPs you can operate on.\\
The current phone's observation: This is a simplified and structured representation of the phone view, providing key information.\\
The previous action and observation : There are the action you just performed and the resulted phone observation. It may be helpful to track your progress.

Solve the user's task with interleaving Observation, Thought, Action steps. \\
Thought can reason about the current situation.\\
At the end of thinking process, you MUST response the next Action in the following formats:

1. APP level Actions:\\
\#start [app-name]\#: This action start an APP specified by app name.
You can ONLY issue the start operation on the following APPs:\\
\{app-string\}

2. Component level Actions:\\
\#click [id]\#: This action clicks on an element with a specific id on the APP page.\\
\#long-click [id]\#: This action long clicks on an element with a specific id on the APP page.\\
\#set-text [id] [text]\# This action set text in a text view element with a specific id on the APP page.\\
Note that the UI elements with 'clickable' or 'long-clickable' properties can be issued with \#click\#, while the elements with 'EditText' can be issued with \#set-text\# action.

3. System level Actions:\\
\#swipe-up\#: Scroll up the screen.\\
\#swipe-down\#: Scroll down the screen.\\
\#swipe-left\#: Swipe left the screen.\\
\#swipe-right\#: Swipe right the screen.\\
\#press-back\#: Navigate to the previously viewed page.\\
\#press-enter\#: Press enter.

4. Completion Action:\\
\#finish [answer]\#: Issue this action when you believe the task is complete. If the objective is to find a text-based answer, provide the answer in the bracket. If you believe the task is impossible to complete, provide the answer as "N/A" in the bracket.

------

Observation is the simplified and structured text representation of APP view.

To be successful, it is very important to follow the following rules:\\
1. You MUST only issue ONE next action in each thinking process.\\
2. Generate the action in the correct format. Always put the action inside a pair of \#. For example, \#click [node3]\#.\\
3. Issue finish action when you think you have achieved the objective.\\
4. Today is \{date\}, which might be useful for you to complete the task.
\end{tcolorbox}

\subsection{Reflection Prompt}
\label{sec:reflect_prompt}
\begin{tcolorbox}[title=Reflection Prompt, breakable]
You are an advanced reasoning agent that can improve based on self reflection. You will be given a previous reasoning trial in which you were given access to operate an Android phone environment with human-like actions including click and type text on the phone screen, and a task instruction to complete. You were unsuccessful in completing the task either because you made the wrong action decisions, or you used up your set number of reasoning steps. In a few sentences, Diagnose a possible reason for failure and devise a new, concise, high level plan that aims to mitigate the same failure. Use complete sentences.
\end{tcolorbox}

\subsection{Reward Prompt}
\label{sec:reward_prompt}
\begin{tcolorbox}[title=Reward Prompt, breakable]
You can access to the actions and phone states at some steps during executing a specific task on a phone. Check if the given phone states and actions indicate the achievement of a goal. The phone state is represented as structured texts, with each entry denoting a UI component along with its content and function description. 

The goal is 
\{goal\}, 

the actions and states at some steps are:
\{traj\}

Please check if the above trajectory indicate the achievement of the goal: {goal}.
Only output 'Yes' or 'No', no other words.
\end{tcolorbox}

\subsection{Query Template}
\label{sec:query_template}
\textbf{Queries for single-APP functionalities extraction}

\begin{tcolorbox}[title=Queries for single-APP, breakable]
how to use \{app\_name\}\\
\{app\_name\} usage instructions\\
\{app\_name\} quick start guides\\
\{app\_name\} cheat sheets\\
\{app\_name\} productivity guides\\
use \{app\_name\} step-by-step\\
tips and tricks for \{app\_name\}\\
\{app\_name\} for beginners\\
\{app\_name\} tutorial\\
getting started with \{app\_name\}\\
introduction to \{app\_name\}
\end{tcolorbox}

\noindent \textbf{Queries for cross-APP functionalities extraction}

\begin{tcolorbox}[title=Queries for cross-APP, breakable]
\{app\_name1\} and \{app\_name2\} collaboration features\\
How to use  \{app\_name1\} and \{app\_name2\} together for tasks\\
Integration between  \{app\_name1\} and \{app\_name2\} for productivity\\
Collaborative task management with  \{app\_name1\} and \{app\_name2\}\\
\{app\_name1\} and \{app\_name2\} integration for work and productivity\\
Productivity tips with  \{app\_name1\} and \{app\_name2\}
\end{tcolorbox}

\subsection{Functionality-to-Instruction Prompt}
\label{sec:f2i_prompt}
\begin{tcolorbox}[title=Functionality-to-Instruction Prompt, breakable]
You are a smart task creator for a smartphone intelligent assistant. Given the features description of the \{app\} APP, your goal is to generate clear and practical tasks that the assistant can assist people with while they use \{app\} on their phone in their daily lives. These tasks should encompass a wide range of possible instructions and questions that may arise when using \{app\} APP.

For example, for the Gmail APP, potential task instructions could include:\\
Compose an email with the subject <email subject> and the message content <email content> to be sent to <email address> using Gmail., \\
Send the first draft email., \\
Open the latest email from <email address> in Gmail., \\
Open Gmail settings., \\
Turn off notifications for Gmail., \\
Star the latest email from <email address> in Gmail., \\
Delete the latest email from <email address> in Gmail., \\
etc., where the placeholders surrounded with angle brackets '<' and '>' should be automated generated and not be filled with specific content.

The \{app\} APP's feature description is: 
\{feature\}

Your task is to generate as many of these tasks as possible for the \{app\} app. Ensure that these instructions are clear and will not lead to any misunderstanding so that the assitant can successfully execute them.
Your response should be a list of comma separated task instructions, where each instruction should be presented in one sentence.
\end{tcolorbox}

%% file: kdd2024.bbl

\begin{thebibliography}{40}


\ifx \showCODEN    \undefined \def \showCODEN     #1{\unskip}     \fi
\ifx \showDOI      \undefined \def \showDOI       #1{#1}\fi
\ifx \showISBNx    \undefined \def \showISBNx     #1{\unskip}     \fi
\ifx \showISBNxiii \undefined \def \showISBNxiii  #1{\unskip}     \fi
\ifx \showISSN     \undefined \def \showISSN      #1{\unskip}     \fi
\ifx \showLCCN     \undefined \def \showLCCN      #1{\unskip}     \fi
\ifx \shownote     \undefined \def \shownote      #1{#1}          \fi
\ifx \showarticletitle \undefined \def \showarticletitle #1{#1}   \fi
\ifx \showURL      \undefined \def \showURL       {\relax}        \fi
\providecommand\bibfield[2]{#2}
\providecommand\bibinfo[2]{#2}
\providecommand\natexlab[1]{#1}
\providecommand\showeprint[2][]{arXiv:#2}

\bibitem[Ahn et~al\mbox{.}(2022)]%
        {ahn2022can}
\bibfield{author}{\bibinfo{person}{Michael Ahn}, \bibinfo{person}{Anthony Brohan}, \bibinfo{person}{Noah Brown}, \bibinfo{person}{Yevgen Chebotar}, \bibinfo{person}{Omar Cortes}, \bibinfo{person}{Byron David}, \bibinfo{person}{Chelsea Finn}, \bibinfo{person}{Chuyuan Fu}, \bibinfo{person}{Keerthana Gopalakrishnan}, \bibinfo{person}{Karol Hausman}, {et~al\mbox{.}}} \bibinfo{year}{2022}\natexlab{}.
\newblock \showarticletitle{Do as i can, not as i say: Grounding language in robotic affordances}.
\newblock \bibinfo{journal}{\emph{arXiv preprint arXiv:2204.01691}} (\bibinfo{year}{2022}).
\newblock


\bibitem[Auer et~al\mbox{.}(2002)]%
        {auer2002finite}
\bibfield{author}{\bibinfo{person}{Peter Auer}, \bibinfo{person}{Nicolo Cesa-Bianchi}, {and} \bibinfo{person}{Paul Fischer}.} \bibinfo{year}{2002}\natexlab{}.
\newblock \showarticletitle{Finite-time analysis of the multiarmed bandit problem}.
\newblock \bibinfo{journal}{\emph{Machine learning}}  \bibinfo{volume}{47} (\bibinfo{year}{2002}), \bibinfo{pages}{235--256}.
\newblock


\bibitem[Austin et~al\mbox{.}(2021)]%
        {austin2021program}
\bibfield{author}{\bibinfo{person}{Jacob Austin}, \bibinfo{person}{Augustus Odena}, \bibinfo{person}{Maxwell Nye}, \bibinfo{person}{Maarten Bosma}, \bibinfo{person}{Henryk Michalewski}, \bibinfo{person}{David Dohan}, \bibinfo{person}{Ellen Jiang}, \bibinfo{person}{Carrie Cai}, \bibinfo{person}{Michael Terry}, \bibinfo{person}{Quoc Le}, {et~al\mbox{.}}} \bibinfo{year}{2021}\natexlab{}.
\newblock \showarticletitle{Program synthesis with large language models}.
\newblock \bibinfo{journal}{\emph{arXiv preprint arXiv:2108.07732}} (\bibinfo{year}{2021}).
\newblock


\bibitem[Carta et~al\mbox{.}(2023)]%
        {carta2023grounding}
\bibfield{author}{\bibinfo{person}{Thomas Carta}, \bibinfo{person}{Cl{\'e}ment Romac}, \bibinfo{person}{Thomas Wolf}, \bibinfo{person}{Sylvain Lamprier}, \bibinfo{person}{Olivier Sigaud}, {and} \bibinfo{person}{Pierre-Yves Oudeyer}.} \bibinfo{year}{2023}\natexlab{}.
\newblock \showarticletitle{Grounding large language models in interactive environments with online reinforcement learning}.
\newblock \bibinfo{journal}{\emph{arXiv preprint arXiv:2302.02662}} (\bibinfo{year}{2023}).
\newblock


\bibitem[Cohen et~al\mbox{.}(2009)]%
        {cohen2009pearson}
\bibfield{author}{\bibinfo{person}{Israel Cohen}, \bibinfo{person}{Yiteng Huang}, \bibinfo{person}{Jingdong Chen}, \bibinfo{person}{Jacob Benesty}, \bibinfo{person}{Jacob Benesty}, \bibinfo{person}{Jingdong Chen}, \bibinfo{person}{Yiteng Huang}, {and} \bibinfo{person}{Israel Cohen}.} \bibinfo{year}{2009}\natexlab{}.
\newblock \showarticletitle{Pearson correlation coefficient}.
\newblock \bibinfo{journal}{\emph{Noise reduction in speech processing}} (\bibinfo{year}{2009}), \bibinfo{pages}{1--4}.
\newblock


\bibitem[Crispino et~al\mbox{.}(2023)]%
        {crispino2023agent}
\bibfield{author}{\bibinfo{person}{Nicholas Crispino}, \bibinfo{person}{Kyle Montgomery}, \bibinfo{person}{Fankun Zeng}, \bibinfo{person}{Dawn Song}, {and} \bibinfo{person}{Chenguang Wang}.} \bibinfo{year}{2023}\natexlab{}.
\newblock \showarticletitle{Agent Instructs Large Language Models to be General Zero-Shot Reasoners}.
\newblock \bibinfo{journal}{\emph{arXiv preprint arXiv:2310.03710}} (\bibinfo{year}{2023}).
\newblock


\bibitem[Deng et~al\mbox{.}(2023)]%
        {deng2023mind2web}
\bibfield{author}{\bibinfo{person}{Xiang Deng}, \bibinfo{person}{Yu Gu}, \bibinfo{person}{Boyuan Zheng}, \bibinfo{person}{Shijie Chen}, \bibinfo{person}{Samuel Stevens}, \bibinfo{person}{Boshi Wang}, \bibinfo{person}{Huan Sun}, {and} \bibinfo{person}{Yu Su}.} \bibinfo{year}{2023}\natexlab{}.
\newblock \showarticletitle{Mind2Web: Towards a Generalist Agent for the Web}.
\newblock \bibinfo{journal}{\emph{arXiv preprint arXiv:2306.06070}} (\bibinfo{year}{2023}).
\newblock


\bibitem[Gunasekaran and Bargavi(2015)]%
        {gunasekaran2015survey}
\bibfield{author}{\bibinfo{person}{S Gunasekaran} {and} \bibinfo{person}{V Bargavi}.} \bibinfo{year}{2015}\natexlab{}.
\newblock \showarticletitle{Survey on automation testing tools for mobile applications}.
\newblock \bibinfo{journal}{\emph{International Journal of Advanced Engineering Research and Science}} \bibinfo{volume}{2}, \bibinfo{number}{11} (\bibinfo{year}{2015}), \bibinfo{pages}{2349--6495}.
\newblock


\bibitem[Hallak et~al\mbox{.}(2015)]%
        {hallak2015contextual}
\bibfield{author}{\bibinfo{person}{Assaf Hallak}, \bibinfo{person}{Dotan Di~Castro}, {and} \bibinfo{person}{Shie Mannor}.} \bibinfo{year}{2015}\natexlab{}.
\newblock \showarticletitle{Contextual markov decision processes}.
\newblock \bibinfo{journal}{\emph{arXiv preprint arXiv:1502.02259}} (\bibinfo{year}{2015}).
\newblock


\bibitem[Huang et~al\mbox{.}(2023)]%
        {huang2023large}
\bibfield{author}{\bibinfo{person}{Jie Huang}, \bibinfo{person}{Xinyun Chen}, \bibinfo{person}{Swaroop Mishra}, \bibinfo{person}{Huaixiu~Steven Zheng}, \bibinfo{person}{Adams~Wei Yu}, \bibinfo{person}{Xinying Song}, {and} \bibinfo{person}{Denny Zhou}.} \bibinfo{year}{2023}\natexlab{}.
\newblock \showarticletitle{Large language models cannot self-correct reasoning yet}.
\newblock \bibinfo{journal}{\emph{arXiv preprint arXiv:2310.01798}} (\bibinfo{year}{2023}).
\newblock


\bibitem[Huang et~al\mbox{.}(2022)]%
        {huang2022language}
\bibfield{author}{\bibinfo{person}{Wenlong Huang}, \bibinfo{person}{Pieter Abbeel}, \bibinfo{person}{Deepak Pathak}, {and} \bibinfo{person}{Igor Mordatch}.} \bibinfo{year}{2022}\natexlab{}.
\newblock \showarticletitle{Language models as zero-shot planners: Extracting actionable knowledge for embodied agents}. In \bibinfo{booktitle}{\emph{International Conference on Machine Learning}}. PMLR, \bibinfo{pages}{9118--9147}.
\newblock


\bibitem[Kwon et~al\mbox{.}(2023)]%
        {kwon2023reward}
\bibfield{author}{\bibinfo{person}{Minae Kwon}, \bibinfo{person}{Sang~Michael Xie}, \bibinfo{person}{Kalesha Bullard}, {and} \bibinfo{person}{Dorsa Sadigh}.} \bibinfo{year}{2023}\natexlab{}.
\newblock \showarticletitle{Reward design with language models}.
\newblock \bibinfo{journal}{\emph{arXiv preprint arXiv:2303.00001}} (\bibinfo{year}{2023}).
\newblock


\bibitem[Li et~al\mbox{.}(2023)]%
        {li2023can}
\bibfield{author}{\bibinfo{person}{Jinyang Li}, \bibinfo{person}{Binyuan Hui}, \bibinfo{person}{Ge Qu}, \bibinfo{person}{Binhua Li}, \bibinfo{person}{Jiaxi Yang}, \bibinfo{person}{Bowen Li}, \bibinfo{person}{Bailin Wang}, \bibinfo{person}{Bowen Qin}, \bibinfo{person}{Rongyu Cao}, \bibinfo{person}{Ruiying Geng}, {et~al\mbox{.}}} \bibinfo{year}{2023}\natexlab{}.
\newblock \showarticletitle{Can llm already serve as a database interface? a big bench for large-scale database grounded text-to-sqls}.
\newblock \bibinfo{journal}{\emph{arXiv preprint arXiv:2305.03111}} (\bibinfo{year}{2023}).
\newblock


\bibitem[Li et~al\mbox{.}(2024)]%
        {li2024personal}
\bibfield{author}{\bibinfo{person}{Yuanchun Li}, \bibinfo{person}{Hao Wen}, \bibinfo{person}{Weijun Wang}, \bibinfo{person}{Xiangyu Li}, \bibinfo{person}{Yizhen Yuan}, \bibinfo{person}{Guohong Liu}, \bibinfo{person}{Jiacheng Liu}, \bibinfo{person}{Wenxing Xu}, \bibinfo{person}{Xiang Wang}, \bibinfo{person}{Yi Sun}, {et~al\mbox{.}}} \bibinfo{year}{2024}\natexlab{}.
\newblock \showarticletitle{Personal LLM Agents: Insights and Survey about the Capability, Efficiency and Security}.
\newblock \bibinfo{journal}{\emph{arXiv preprint arXiv:2401.05459}} (\bibinfo{year}{2024}).
\newblock


\bibitem[Liu et~al\mbox{.}(2023)]%
        {liu2023agentbench}
\bibfield{author}{\bibinfo{person}{Xiao Liu}, \bibinfo{person}{Hao Yu}, \bibinfo{person}{Hanchen Zhang}, \bibinfo{person}{Yifan Xu}, \bibinfo{person}{Xuanyu Lei}, \bibinfo{person}{Hanyu Lai}, \bibinfo{person}{Yu Gu}, \bibinfo{person}{Hangliang Ding}, \bibinfo{person}{Kaiwen Men}, \bibinfo{person}{Kejuan Yang}, {et~al\mbox{.}}} \bibinfo{year}{2023}\natexlab{}.
\newblock \showarticletitle{Agentbench: Evaluating llms as agents}.
\newblock \bibinfo{journal}{\emph{arXiv preprint arXiv:2308.03688}} (\bibinfo{year}{2023}).
\newblock


\bibitem[Ma et~al\mbox{.}(2023)]%
        {ma2023laser}
\bibfield{author}{\bibinfo{person}{Kaixin Ma}, \bibinfo{person}{Hongming Zhang}, \bibinfo{person}{Hongwei Wang}, \bibinfo{person}{Xiaoman Pan}, {and} \bibinfo{person}{Dong Yu}.} \bibinfo{year}{2023}\natexlab{}.
\newblock \showarticletitle{LASER: LLM Agent with State-Space Exploration for Web Navigation}.
\newblock \bibinfo{journal}{\emph{arXiv preprint arXiv:2309.08172}} (\bibinfo{year}{2023}).
\newblock


\bibitem[Mnih et~al\mbox{.}(2013)]%
        {mnih2013playing}
\bibfield{author}{\bibinfo{person}{Volodymyr Mnih}, \bibinfo{person}{Koray Kavukcuoglu}, \bibinfo{person}{David Silver}, \bibinfo{person}{Alex Graves}, \bibinfo{person}{Ioannis Antonoglou}, \bibinfo{person}{Daan Wierstra}, {and} \bibinfo{person}{Martin Riedmiller}.} \bibinfo{year}{2013}\natexlab{}.
\newblock \showarticletitle{Playing atari with deep reinforcement learning}.
\newblock \bibinfo{journal}{\emph{arXiv preprint arXiv:1312.5602}} (\bibinfo{year}{2013}).
\newblock


\bibitem[OpenAI(2023)]%
        {openai2023gpt}
\bibfield{author}{\bibinfo{person}{R OpenAI}.} \bibinfo{year}{2023}\natexlab{}.
\newblock \showarticletitle{GPT-4 technical report}.
\newblock \bibinfo{journal}{\emph{arXiv}} (\bibinfo{year}{2023}), \bibinfo{pages}{2303--08774}.
\newblock


\bibitem[Qian et~al\mbox{.}(2023)]%
        {qian2023communicative}
\bibfield{author}{\bibinfo{person}{Chen Qian}, \bibinfo{person}{Xin Cong}, \bibinfo{person}{Cheng Yang}, \bibinfo{person}{Weize Chen}, \bibinfo{person}{Yusheng Su}, \bibinfo{person}{Juyuan Xu}, \bibinfo{person}{Zhiyuan Liu}, {and} \bibinfo{person}{Maosong Sun}.} \bibinfo{year}{2023}\natexlab{}.
\newblock \showarticletitle{Communicative agents for software development}.
\newblock \bibinfo{journal}{\emph{arXiv preprint arXiv:2307.07924}} (\bibinfo{year}{2023}).
\newblock


\bibitem[Rawles et~al\mbox{.}(2023)]%
        {rawles2023aitw}
\bibfield{author}{\bibinfo{person}{Christopher Rawles}, \bibinfo{person}{Alice Li}, \bibinfo{person}{Daniel Rodriguez}, \bibinfo{person}{Oriana Riva}, {and} \bibinfo{person}{Timothy Lillicrap}.} \bibinfo{year}{2023}\natexlab{}.
\newblock \showarticletitle{Android in the wild: A large-scale dataset for android device control}.
\newblock \bibinfo{journal}{\emph{arXiv preprint arXiv:2307.10088}} (\bibinfo{year}{2023}).
\newblock


\bibitem[Ruan et~al\mbox{.}(2023)]%
        {ruan2023tptu}
\bibfield{author}{\bibinfo{person}{Jingqing Ruan}, \bibinfo{person}{Yihong Chen}, \bibinfo{person}{Bin Zhang}, \bibinfo{person}{Zhiwei Xu}, \bibinfo{person}{Tianpeng Bao}, \bibinfo{person}{Guoqing Du}, \bibinfo{person}{Shiwei Shi}, \bibinfo{person}{Hangyu Mao}, \bibinfo{person}{Xingyu Zeng}, {and} \bibinfo{person}{Rui Zhao}.} \bibinfo{year}{2023}\natexlab{}.
\newblock \showarticletitle{Tptu: Task planning and tool usage of large language model-based ai agents}.
\newblock \bibinfo{journal}{\emph{arXiv preprint arXiv:2308.03427}} (\bibinfo{year}{2023}).
\newblock


\bibitem[Shinn et~al\mbox{.}(2023)]%
        {shinn2023reflexion}
\bibfield{author}{\bibinfo{person}{Noah Shinn}, \bibinfo{person}{Federico Cassano}, \bibinfo{person}{Ashwin Gopinath}, \bibinfo{person}{Karthik~R Narasimhan}, {and} \bibinfo{person}{Shunyu Yao}.} \bibinfo{year}{2023}\natexlab{}.
\newblock \showarticletitle{Reflexion: Language agents with verbal reinforcement learning}. In \bibinfo{booktitle}{\emph{Thirty-seventh Conference on Neural Information Processing Systems}}.
\newblock


\bibitem[Shridhar et~al\mbox{.}(2020)]%
        {shridhar2020alfworld}
\bibfield{author}{\bibinfo{person}{Mohit Shridhar}, \bibinfo{person}{Xingdi Yuan}, \bibinfo{person}{Marc-Alexandre C{\^o}t{\'e}}, \bibinfo{person}{Yonatan Bisk}, \bibinfo{person}{Adam Trischler}, {and} \bibinfo{person}{Matthew Hausknecht}.} \bibinfo{year}{2020}\natexlab{}.
\newblock \showarticletitle{Alfworld: Aligning text and embodied environments for interactive learning}.
\newblock \bibinfo{journal}{\emph{arXiv preprint arXiv:2010.03768}} (\bibinfo{year}{2020}).
\newblock


\bibitem[Sutton et~al\mbox{.}(1999)]%
        {sutton1999reinforcement}
\bibfield{author}{\bibinfo{person}{Richard~S Sutton}, \bibinfo{person}{Andrew~G Barto}, {et~al\mbox{.}}} \bibinfo{year}{1999}\natexlab{}.
\newblock \showarticletitle{Reinforcement learning}.
\newblock \bibinfo{journal}{\emph{Journal of Cognitive Neuroscience}} \bibinfo{volume}{11}, \bibinfo{number}{1} (\bibinfo{year}{1999}), \bibinfo{pages}{126--134}.
\newblock


\bibitem[Touvron et~al\mbox{.}(2023)]%
        {touvron2023llama}
\bibfield{author}{\bibinfo{person}{Hugo Touvron}, \bibinfo{person}{Louis Martin}, \bibinfo{person}{Kevin Stone}, \bibinfo{person}{Peter Albert}, \bibinfo{person}{Amjad Almahairi}, \bibinfo{person}{Yasmine Babaei}, \bibinfo{person}{Nikolay Bashlykov}, \bibinfo{person}{Soumya Batra}, \bibinfo{person}{Prajjwal Bhargava}, \bibinfo{person}{Shruti Bhosale}, {et~al\mbox{.}}} \bibinfo{year}{2023}\natexlab{}.
\newblock \showarticletitle{Llama 2: Open foundation and fine-tuned chat models}.
\newblock \bibinfo{journal}{\emph{arXiv preprint arXiv:2307.09288}} (\bibinfo{year}{2023}).
\newblock


\bibitem[Toyama et~al\mbox{.}(2021)]%
        {ToyamaEtAl2021AndroidEnv}
\bibfield{author}{\bibinfo{person}{Daniel Toyama}, \bibinfo{person}{Philippe Hamel}, \bibinfo{person}{Anita Gergely}, \bibinfo{person}{Gheorghe Comanici}, \bibinfo{person}{Amelia Glaese}, \bibinfo{person}{Zafarali Ahmed}, \bibinfo{person}{Tyler Jackson}, \bibinfo{person}{Shibl Mourad}, {and} \bibinfo{person}{Doina Precup}.} \bibinfo{year}{2021}\natexlab{}.
\newblock \showarticletitle{{AndroidEnv}: A Reinforcement Learning Platform for Android}.
\newblock   \bibinfo{volume}{abs/2105.13231} (\bibinfo{year}{2021}).
\newblock
\showeprint[arxiv]{2105.13231}~[cs.LG]
\urldef\tempurl%
\url{http://arxiv.org/abs/2105.13231}
\showURL{%
\tempurl}


\bibitem[Wang et~al\mbox{.}(2023c)]%
        {wang2023can}
\bibfield{author}{\bibinfo{person}{Boshi Wang}, \bibinfo{person}{Xiang Yue}, {and} \bibinfo{person}{Huan Sun}.} \bibinfo{year}{2023}\natexlab{c}.
\newblock \showarticletitle{Can ChatGPT Defend its Belief in Truth? Evaluating LLM Reasoning via Debate}. In \bibinfo{booktitle}{\emph{Findings of the Association for Computational Linguistics: EMNLP 2023}}. \bibinfo{pages}{11865--11881}.
\newblock


\bibitem[Wang et~al\mbox{.}(2023b)]%
        {wang2023voyager}
\bibfield{author}{\bibinfo{person}{Guanzhi Wang}, \bibinfo{person}{Yuqi Xie}, \bibinfo{person}{Yunfan Jiang}, \bibinfo{person}{Ajay Mandlekar}, \bibinfo{person}{Chaowei Xiao}, \bibinfo{person}{Yuke Zhu}, \bibinfo{person}{Linxi Fan}, {and} \bibinfo{person}{Anima Anandkumar}.} \bibinfo{year}{2023}\natexlab{b}.
\newblock \showarticletitle{Voyager: An open-ended embodied agent with large language models}.
\newblock \bibinfo{journal}{\emph{arXiv preprint arXiv:2305.16291}} (\bibinfo{year}{2023}).
\newblock


\bibitem[Wang et~al\mbox{.}(2023a)]%
        {wang2023survey}
\bibfield{author}{\bibinfo{person}{Lei Wang}, \bibinfo{person}{Chen Ma}, \bibinfo{person}{Xueyang Feng}, \bibinfo{person}{Zeyu Zhang}, \bibinfo{person}{Hao Yang}, \bibinfo{person}{Jingsen Zhang}, \bibinfo{person}{Zhiyuan Chen}, \bibinfo{person}{Jiakai Tang}, \bibinfo{person}{Xu Chen}, \bibinfo{person}{Yankai Lin}, {et~al\mbox{.}}} \bibinfo{year}{2023}\natexlab{a}.
\newblock \showarticletitle{A survey on large language model based autonomous agents}.
\newblock \bibinfo{journal}{\emph{arXiv preprint arXiv:2308.11432}} (\bibinfo{year}{2023}).
\newblock


\bibitem[Wen et~al\mbox{.}(2022)]%
        {improving_ucb}
\bibfield{author}{\bibinfo{person}{Yingpeng Wen}, \bibinfo{person}{Qinliang Su}, \bibinfo{person}{Minghua Shen}, {and} \bibinfo{person}{Nong Xiao}.} \bibinfo{year}{2022}\natexlab{}.
\newblock \showarticletitle{Improving the exploration efficiency of DQNs via the confidence bound methods}.
\newblock \bibinfo{journal}{\emph{Applied Intelligence}} (\bibinfo{year}{2022}), \bibinfo{pages}{1--15}.
\newblock


\bibitem[Wu et~al\mbox{.}(2023)]%
        {wu2023smartplay}
\bibfield{author}{\bibinfo{person}{Yue Wu}, \bibinfo{person}{Xuan Tang}, \bibinfo{person}{Tom~M Mitchell}, {and} \bibinfo{person}{Yuanzhi Li}.} \bibinfo{year}{2023}\natexlab{}.
\newblock \showarticletitle{SmartPlay: A Benchmark for LLMs as Intelligent Agents}.
\newblock \bibinfo{journal}{\emph{arXiv preprint arXiv:2310.01557}} (\bibinfo{year}{2023}).
\newblock


\bibitem[Xi et~al\mbox{.}(2023)]%
        {xi2023rise}
\bibfield{author}{\bibinfo{person}{Zhiheng Xi}, \bibinfo{person}{Wenxiang Chen}, \bibinfo{person}{Xin Guo}, \bibinfo{person}{Wei He}, \bibinfo{person}{Yiwen Ding}, \bibinfo{person}{Boyang Hong}, \bibinfo{person}{Ming Zhang}, \bibinfo{person}{Junzhe Wang}, \bibinfo{person}{Senjie Jin}, \bibinfo{person}{Enyu Zhou}, {et~al\mbox{.}}} \bibinfo{year}{2023}\natexlab{}.
\newblock \showarticletitle{The rise and potential of large language model based agents: A survey}.
\newblock \bibinfo{journal}{\emph{arXiv preprint arXiv:2309.07864}} (\bibinfo{year}{2023}).
\newblock


\bibitem[Xu et~al\mbox{.}(2023)]%
        {xu2023wizardlm}
\bibfield{author}{\bibinfo{person}{Can Xu}, \bibinfo{person}{Qingfeng Sun}, \bibinfo{person}{Kai Zheng}, \bibinfo{person}{Xiubo Geng}, \bibinfo{person}{Pu Zhao}, \bibinfo{person}{Jiazhan Feng}, \bibinfo{person}{Chongyang Tao}, {and} \bibinfo{person}{Daxin Jiang}.} \bibinfo{year}{2023}\natexlab{}.
\newblock \showarticletitle{Wizardlm: Empowering large language models to follow complex instructions}.
\newblock \bibinfo{journal}{\emph{arXiv preprint arXiv:2304.12244}} (\bibinfo{year}{2023}).
\newblock


\bibitem[Yamada et~al\mbox{.}(2023)]%
        {yamada2023evaluating}
\bibfield{author}{\bibinfo{person}{Yutaro Yamada}, \bibinfo{person}{Yihan Bao}, \bibinfo{person}{Andrew~K Lampinen}, \bibinfo{person}{Jungo Kasai}, {and} \bibinfo{person}{Ilker Yildirim}.} \bibinfo{year}{2023}\natexlab{}.
\newblock \showarticletitle{Evaluating Spatial Understanding of Large Language Models}.
\newblock \bibinfo{journal}{\emph{arXiv preprint arXiv:2310.14540}} (\bibinfo{year}{2023}).
\newblock


\bibitem[Yan et~al\mbox{.}(2023)]%
        {yan2023gpt}
\bibfield{author}{\bibinfo{person}{An Yan}, \bibinfo{person}{Zhengyuan Yang}, \bibinfo{person}{Wanrong Zhu}, \bibinfo{person}{Kevin Lin}, \bibinfo{person}{Linjie Li}, \bibinfo{person}{Jianfeng Wang}, \bibinfo{person}{Jianwei Yang}, \bibinfo{person}{Yiwu Zhong}, \bibinfo{person}{Julian McAuley}, \bibinfo{person}{Jianfeng Gao}, {et~al\mbox{.}}} \bibinfo{year}{2023}\natexlab{}.
\newblock \showarticletitle{Gpt-4v in wonderland: Large multimodal models for zero-shot smartphone gui navigation}.
\newblock \bibinfo{journal}{\emph{arXiv preprint arXiv:2311.07562}} (\bibinfo{year}{2023}).
\newblock


\bibitem[Yang et~al\mbox{.}(2023)]%
        {yang2023appagent}
\bibfield{author}{\bibinfo{person}{Zhao Yang}, \bibinfo{person}{Jiaxuan Liu}, \bibinfo{person}{Yucheng Han}, \bibinfo{person}{Xin Chen}, \bibinfo{person}{Zebiao Huang}, \bibinfo{person}{Bin Fu}, {and} \bibinfo{person}{Gang Yu}.} \bibinfo{year}{2023}\natexlab{}.
\newblock \showarticletitle{Appagent: Multimodal agents as smartphone users}.
\newblock \bibinfo{journal}{\emph{arXiv preprint arXiv:2312.13771}} (\bibinfo{year}{2023}).
\newblock


\bibitem[Yang et~al\mbox{.}(2018)]%
        {yang2018hotpotqa}
\bibfield{author}{\bibinfo{person}{Zhilin Yang}, \bibinfo{person}{Peng Qi}, \bibinfo{person}{Saizheng Zhang}, \bibinfo{person}{Yoshua Bengio}, \bibinfo{person}{William~W Cohen}, \bibinfo{person}{Ruslan Salakhutdinov}, {and} \bibinfo{person}{Christopher~D Manning}.} \bibinfo{year}{2018}\natexlab{}.
\newblock \showarticletitle{HotpotQA: A dataset for diverse, explainable multi-hop question answering}.
\newblock \bibinfo{journal}{\emph{arXiv preprint arXiv:1809.09600}} (\bibinfo{year}{2018}).
\newblock


\bibitem[Yao et~al\mbox{.}(2022)]%
        {yao2022react}
\bibfield{author}{\bibinfo{person}{Shunyu Yao}, \bibinfo{person}{Jeffrey Zhao}, \bibinfo{person}{Dian Yu}, \bibinfo{person}{Nan Du}, \bibinfo{person}{Izhak Shafran}, \bibinfo{person}{Karthik Narasimhan}, {and} \bibinfo{person}{Yuan Cao}.} \bibinfo{year}{2022}\natexlab{}.
\newblock \showarticletitle{React: Synergizing reasoning and acting in language models}.
\newblock \bibinfo{journal}{\emph{arXiv preprint arXiv:2210.03629}} (\bibinfo{year}{2022}).
\newblock


\bibitem[Yuan et~al\mbox{.}(2023)]%
        {yuan2023plan4mc}
\bibfield{author}{\bibinfo{person}{Haoqi Yuan}, \bibinfo{person}{Chi Zhang}, \bibinfo{person}{Hongcheng Wang}, \bibinfo{person}{Feiyang Xie}, \bibinfo{person}{Penglin Cai}, \bibinfo{person}{Hao Dong}, {and} \bibinfo{person}{Zongqing Lu}.} \bibinfo{year}{2023}\natexlab{}.
\newblock \showarticletitle{Plan4mc: Skill reinforcement learning and planning for open-world minecraft tasks}.
\newblock \bibinfo{journal}{\emph{arXiv preprint arXiv:2303.16563}} (\bibinfo{year}{2023}).
\newblock


\bibitem[Zhou et~al\mbox{.}(2023)]%
        {zhou2023webarena}
\bibfield{author}{\bibinfo{person}{Shuyan Zhou}, \bibinfo{person}{Frank~F Xu}, \bibinfo{person}{Hao Zhu}, \bibinfo{person}{Xuhui Zhou}, \bibinfo{person}{Robert Lo}, \bibinfo{person}{Abishek Sridhar}, \bibinfo{person}{Xianyi Cheng}, \bibinfo{person}{Yonatan Bisk}, \bibinfo{person}{Daniel Fried}, \bibinfo{person}{Uri Alon}, {et~al\mbox{.}}} \bibinfo{year}{2023}\natexlab{}.
\newblock \showarticletitle{Webarena: A realistic web environment for building autonomous agents}.
\newblock \bibinfo{journal}{\emph{arXiv preprint arXiv:2307.13854}} (\bibinfo{year}{2023}).
\newblock


\end{thebibliography}
